\documentclass[10pt,journal,compsoc]{IEEEtran}
\ifCLASSOPTIONcompsoc
  \usepackage[nocompress]{cite}
\else
  \usepackage{cite}
\fi
\ifCLASSINFOpdf
  \usepackage[pdftex]{graphicx}
\else
\fi
\usepackage{amsmath}
\usepackage{algorithmic}

\usepackage{array}

\ifCLASSOPTIONcompsoc
  \usepackage[caption=false,font=footnotesize,labelfont=sf,textfont=sf]{subfig}
\else
  \usepackage[caption=false,font=footnotesize]{subfig}
\fi
\usepackage{fixltx2e}

\usepackage{stfloats}
\usepackage{url}

\usepackage{amsmath,amsfonts}
\usepackage{array}
\usepackage{booktabs}
\usepackage{multirow}
\usepackage{hyperref}
\hypersetup{colorlinks,linkcolor={red},citecolor={cyan},urlcolor={magenta}} 
\usepackage{breakcites}
\usepackage[flushleft]{threeparttable}
\usepackage[misc,geometry]{ifsym}
\usepackage[table]{xcolor}
\definecolor{LightCyan}{rgb}{0.88,1,1}
\definecolor{Gray}{gray}{0.8999}
\definecolor{LightGray}{gray}{0.9486}

\usepackage{ragged2e}

\usepackage[outline]{contour}
\contourlength{0.07pt}
\contournumber{10}%

\begin{document}
%
\title{Semantic-Aligned Matching for Enhanced DETR Convergence and Multi-Scale Feature Fusion}
%
%
%
%

\author{Gongjie Zhang$^\dagger$, Zhipeng Luo$^\dagger$, Jiaxing Huang, Shijian Lu\,$^{\text{\Letter}}$, and Eric P. Xing
\IEEEcompsocitemizethanks{\IEEEcompsocthanksitem Gongjie Zhang, Zhipeng Luo, Jiaxing Huang, and Shijian Lu are with the School of Computer Science and Engineering, Nanyang Technological University, Singapore.\protect
\IEEEcompsocthanksitem Eric P. Xing is with the School of Computer Science, Carnegie Mellon University. He also serves as the president of Mohamed bin Zayed University of Artificial Intelligence.\protect
\IEEEcompsocthanksitem E-mail: gjz@ieee.org{\tiny\,}, Zhipeng001@e.ntu.edu.sg{\tiny\,}, Shijian.Lu@ntu.edu.sg{\tiny\,}, Jiaxing.Huang@ntu.edu.sg, Eric.Xing@mbzuai.ac.ae{\tiny\,}.
\IEEEcompsocthanksitem $\dagger$ denotes equal contribution; \;${\text{\Letter}}$ denotes corresponding author.}
\thanks{Preprint version. All rights reserved by the authors.}}

%
%

\markboth{Zhang \textit{et al.}:\; Semantic-Aligned Matching for Enhanced DETR Convergence and Multi-Scale Feature Fusion}%
{Shell \MakeLowercase{\textit{et al.}}: Bare Demo of IEEEtran.cls for Computer Society Journals}
%



\IEEEtitleabstractindextext{%
\begin{abstract}
    \RaggedRight
    The recently proposed DEtection TRansformer (DETR) has established a fully end-to-end paradigm for object detection. However, DETR suffers from slow training convergence, which hinders its applicability to various detection tasks. We observe that DETR's slow convergence is largely attributed to the difficulty in matching object queries to relevant regions due to the unaligned semantics between object queries and encoded image features. With this observation, we design \textit{Semantic-Aligned-Matching\;DETR++ (SAM-DETR++)} to accelerate DETR's convergence and improve detection performance. The core of SAM-DETR++ is a plug-and-play module that projects object queries and encoded image features into the same feature embedding space, where each object query can be easily matched to relevant regions with similar semantics. Besides, SAM-DETR++ searches for multiple representative keypoints and exploits their features for semantic-aligned matching with enhanced representation capacity. Furthermore, SAM-DETR++ can effectively fuse multi-scale features in a coarse-to-fine manner on the basis of the designed semantic-aligned matching. Extensive experiments show that the proposed SAM-DETR++ achieves superior convergence speed and competitive detection accuracy. Additionally, as a plug-and-play method, SAM-DETR++ can complement existing DETR convergence solutions with even better performance, achieving 44.8\%\,AP with merely 12 training epochs and 49.1\%\,AP with 50 training epochs on COCO val\,2017\\with ResNet-50. Codes are available at {\tiny\,}\href{https://github.com/ZhangGongjie/SAM-DETR}{https://github.com/ZhangGongjie/SAM-DETR}{\tiny\,}.
\end{abstract}

}

\maketitle

\IEEEdisplaynontitleabstractindextext

%
\IEEEpeerreviewmaketitle

\IEEEraisesectionheading{\section{Introduction}\label{sec:introduction}}

%
%
%
%
\IEEEPARstart{O}{bject} detection~\cite{Liu2019DeepLF} is a fundamental computer vision task and has experienced remarkable progress with the recent development of deep learning and convolutional neural networks (ConvNets). However, most modern ConvNet-based object detectors (\textit{e.g.}, Faster R-CNN~\cite{FasterRCNN}, YOLO~\cite{YOLO9000}, FCOS~\cite{FCOS}) still heavily rely on a series of hand-crafted components, such as anchors, non-maximum suppression (NMS), rule-based training target assignment, \textit{etc.}, which lead to complex detection pipelines and sub-optimal performance. Recently, the emergence of DEtection TRansformer (DETR)~\cite{DETR} has revolutionized the paradigm for object detection. DETR adopts a simple Transformer encoder-decoder pipeline~\cite{transformer} and removes the need for those hand-crafted components, achieving a fully end-to-end framework for object detection. However, despite its simplicity and promising performance, DETR suffers from severely slow convergence on training, requiring 500 epochs to fully converge on the COCO dataset~\cite{MSCOCO}, while most other ConvNet-based object detectors~\cite{FasterRCNN,YOLO9000,focalloss,FCOS} only requires 12$\rm\sim$36 training epochs instead. DETR's slow convergence significantly increases its training cost and thus hinders the wide application of DETR.

\begin{figure}[t!] 
\begin{center}
   \includegraphics[width=1.0\linewidth]{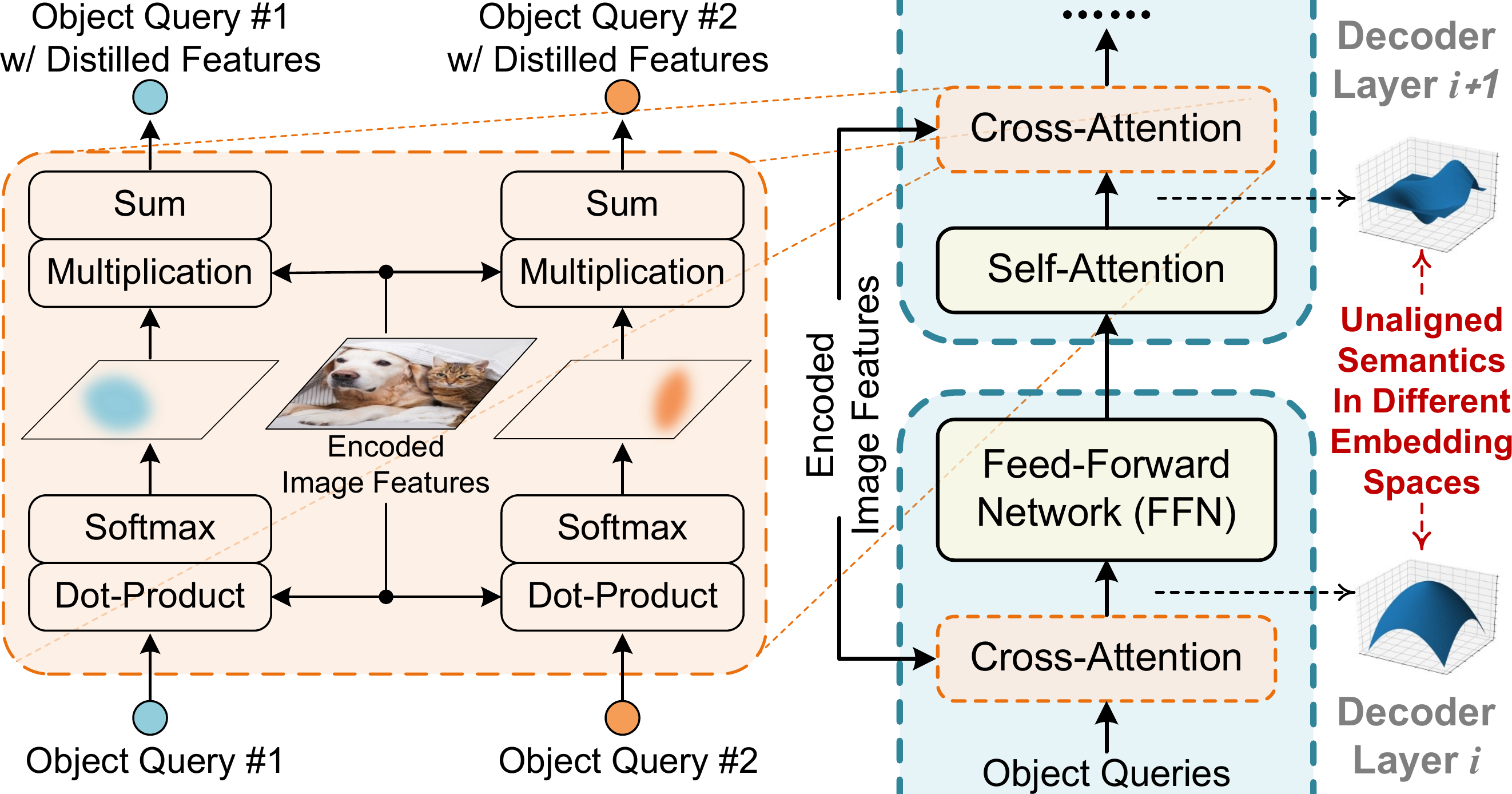}
\end{center}
\vspace{-3.5mm}
   \caption{
   The analysis for the root of DETR's slow convergence. \textbf{\textit{\,Left:}} The cross-attention module in DETR's decoder layers can be interpreted as a `matching and feature distillation' process. Each object query first matches its particular relevant regions in encoded image features via `Dot-Product and Softmax', and then distills instance-level features from the matched regions for subsequent prediction. \textbf{\textit{\,Right:}} However, modules between cross-attentions may project object queries and encoded image features into different feature embedding spaces, leading to the unaligned semantics between them. Such unaligned semantics imposes difficulty in cross-attention's matching process and thus hinders the convergence of Transformer-based object detection frameworks.
   }
\label{fig:fig1}
\end{figure}

DETR uses a set of object queries in its decoder to represent potential objects at different spatial locations. As shown in Fig.\;\ref{fig:fig1}\;(left), in the cross-attention modules of DETR's decoder layers, these object queries interact with the encoded image features through a `matching and feature distillation' process, where each object query first matches its relevant regions in the encoded image features, and then distills corresponding instance-level features from the matched regions. The object queries after distilling relevant features are used to generate instance-level detection predictions as well as to repeat the subsequent `matching and feature distillation' processes for refined predictions. However, as pointed out in \cite{DeformableDETR,ConditionalDETR,SMCA-DETR,AnchorDETR,SAM-DETR}, it is difficult for object queries to learn to match appropriate regions. As illustrated in Fig.\;\ref{fig:fig1}\;(right), we observe that the matching difficulty is largely attributed to the unaligned semantics between object queries and encoded image features. Concretely, the modules between cross-attentions project object queries into different feature embedding spaces, in which object queries have different feature semantics from encoded image features. This leads to the complexity of matching object queries with relevant regions and further DETR's slow training convergence.

An intuitive and promising direction to mitigate the matching difficulty caused by unaligned semantics has been explored in Siamese-based architectures, which adopt identical sub-networks to produce comparable output feature vectors for similarity computation. The effectiveness of Siamese-based architectures has been extensively verified in various matching-involved vision tasks, including object tracking~\cite{Siam-FC,SiamRPN,SiamRPN++,SiamRCNN,TransformerTrack,TransT}, re-identification\cite{chung2017two,zheng2019re,wu2018and,shen2017deep,Shen_2017_ICCV}, and few-shot recognition~\cite{SiameseOneshotImageRecognition,ProtoNet,RelationNetwork,NEURIPS2019_92af93f7,MetaDETR}. In light of the success of Siamese-based architectures in matching-involved tasks, we follow the similar philosophy to address the matching difficulty in the cross-attention module of DETR's decoder.

With these motivations, we propose \textit{Semantic-Aligned-Matching\;DETR++ (SAM-DETR++)} that accelerates the convergence of DETR via a semantic-aligned matching mechanism. Concretely, SAM-DETR++ appends a plug-and-play module ahead of the cross-attention modules in DETR's decoder layers, with which object queries and encoded image features can be projected into the same semantics-aligned feature embedding spaces and thus be matched efficiently. The aligned semantics imposes a strong prior for each object query to focus on those semantically similar regions in encoded image features. SAM-DETR++ also explicitly identifies multiple representative keypoints for each object query and exploits their features for semantic-aligned matching, which can naturally fit into the original multi-head attention mechanism~\cite{transformer} for enhanced representation capacity. In addition, we extend the semantic-aligned matching mechanism to incorporate multi-scale features that are inherently unaligned in feature semantics. This enables SAM-DETR++ to represent objects at different scales in a `divide and conquer' manner and significantly alleviates the representation complexity, yielding faster convergence and improved accuracy. Further, as SAM-DETR++ works as a plug-in to the original DETR~\cite{DETR} with little modification to the rest operations, it can be easily integrated with existing DETR convergence strategies~\cite{SMCA-DETR,DN-DETR} in a complementary manner, boosting detection performance to a greater extent.

\vspace{+0.2mm}
The contributions of this work are summarized below.
\textit{\textbf{(i)}} We propose \textit{SAM-DETR++}, which accelerates DETR's convergence with a plug-and-play module that enables semantic-aligned matching between object queries and encoded image features.
\textit{\textbf{\;(ii)}} We propose to explicitly search for objects' representative keypoints and leverage their features for semantic-aligned matching, which further strengthens the representation capacity of the introduced semantic-aligned matching mechanism.
\textit{\textbf{\;(iii)}} We introduce a multi-scale design into the semantic-aligned matching mechanism to effectively fuse multi-scale features in a coarse-to-fine manner, which enables adaptive representation of objects at different scales and achieves faster convergence as well as superior detection performance.
\textit{\textbf{\;(iv)}\;} Our approach offers a unique perspective in mitigating DETR's slow convergence issue with simply a plug-and-play module, thus can be easily integrated with existing convergence solutions in a complementary manner. Experiments show that with just 12 training epochs, our fully-fledged SAM-DETR++ surpasses the original DETR~\cite{DETR} trained for 500 epochs on the COCO benchmark~\cite{MSCOCO}, and achieves the state-of-the-art performance among DETR-based detectors.

This paper is an extension of our previous paper~\cite{SAM-DETR} published at the \textit{CVPR'\,2022} conference. Compared with its conference version~\cite{SAM-DETR}, this paper incorporates the following new contributions.
\textit{\textbf{\;(i)}} We extend the proposed semantic-aligned matching mechanism to effectively incorporate the multi-scale features that are inherently unaligned in semantics. This enables the adaptive representation of objects of different sizes and and further improves the convergence speed and detection accuracy significantly.
\textit{\textbf{\;(ii)}} We examine the compatibility of SAM-DETR++ with more recently proposed DETR convergence solutions, demonstrating its superior robustness and achieving further improvement in detection accuracy.
\textit{\textbf{\;(iii)}} A minor tweak to remove the dropout in the Transformer is adopted for superior performance at no computational cost.
\textit{\textbf{\;(iv)}} We conduct more comprehensive analysis of SAM-DETR++, including visualization, illustration, and experimentation, for a clearer explanation and a better understanding of our method.

\vspace{0.35mm}
The rest of this paper is organized as follows: Section~\ref{sec:related_work} presents related work; Section~\ref{sec:DETR_Review} briefly reviews the architecture of DETR~\cite{DETR}; Section~\ref{sec:method} describes our proposed method in detail; Section~\ref{sec:experiments} presents experiment results and our analysis; Section~\ref{sec:SAMDETR_conclusion} draws the concluding remarks.

\section{Related Work}   \label{sec:related_work}

\subsection{ConvNet-Based Object Detection}

Most modern object detectors are based on ConvNets and have experienced remarkable progress with the development of deep learning~\cite{Liu2019DeepLF}. Most of these ConvNet-based detectors can be divided into two categories: two-stage and one-stage detectors. Two-stage detectors mainly involve Faster R-CNN~\cite{FasterRCNN} and its extensions~\cite{CascadeRCNN,LibraRCNN,tychsen2018improving,RelationNetworkObjectDetection,CADNet,metarcnn,masktextspotter,FSDetView,fsod,fsdet}, which use a region proposal network (RPN) to first produce region proposals and then perform region-wise predictions over the proposals. These two-stage detectors are generally more accurate and have achieved state-of-the-art performance in various object detection challenges. Differently, one-stage detectors~\cite{SSD,YOLO9000,focalloss,FCOS,FewshotReweighting,RefineDet,RFBNet,m2det,efficientdet,zhou2019objects,incrementalfsdet,PNPDet,ExtremeNet} remove the proposal generation stage and directly predict over the densely placed shape priors, which achieve better inference speed.

Even with promising results, these ConvNet-based object detectors are still not optimal. They perform object detection by defining and solving surrogate regression and classification tasks, and rely on many hand-crafted components, such as non-maximum suppression (NMS), anchors, and rule-based training target assignment. Therefore, these ConvNet-based detectors' architectures are rather complicated, hyper-parameter-intensive, and not fully end-to-end, which leads to sub-optimal detection performance.

\subsection{Transformer-Based Object Detection}

Distinct from those ConvNet-based detectors, the recently proposed DETR~\cite{DETR} has revolutionized the object detection paradigm using a Transformer architecture~\cite{transformer} optimized by a set-based global loss. DETR eliminates the need for many hand-designed components and achieves the first fully end-to-end object detector with competitive performance. However, DETR suffers from extremely slow convergence and needs extra-long training to achieve good performance compared with those ConvNet-based object detectors. A few recent works have been proposed to mitigate this issue. Deformable DETR~\cite{DeformableDETR}, Efficient DETR~\cite{EfficientDETR}, Sparse DETR~\cite{SparseDETR}, and ViDT~\cite{ViDT} replace the original dense attention with sparse attention mechanisms. PnP-DETR~\cite{PnPDETR} proposes a poll-and-pool sampling strategy in its attention mechanism. Besides, Conditional DETR~\cite{ConditionalDETR}, SMCA-DETR~\cite{SMCA-DETR}, Anchor DETR~\cite{AnchorDETR}, and DAB-DETR~\cite{DABDETR} make substantial modifications to the attention mechanism, aiming to add spatial constraints to the original cross-attention to better focus on prominent regions. Furthermore, the recently proposed DN-DETR~\cite{DN-DETR} designs a novel de-noising training strategy to speed up DETR's training procedure, which also achieves very promising results.

In this work, we also aim to improve DETR's convergence and performance, but from a distinctive perspective. Our method does not modify the original attention mechanism in DETR, nor does it change the training strategy for DETR. Our method notices the unaligned semantics between object queries and encoded image features, which causes the difficulty of matching object queries to relevant regions and further DETR's slow convergence. To address this issue, our method only appends a plug-and-play module into the DETR architecture and thus can naturally work with existing convergence solutions for DETR~\cite{SMCA-DETR,DN-DETR} in complementary. Besides, our method can be easily extended to fuse multi-scale features, which further reduces the complexity in representing objects of different sizes, thus accelerating the convergence to a greater extent.

\subsection{Siamese Architectures for Matching-Involved Tasks}

Matching is a common task in computer vision, of which the core idea is to predict the similarity between a pair of inputs. The concept of matching is commonly referred to in contrastive tasks, like face recognition~\cite{FaceNet,song2019occlusion}, object tracking~\cite{Siam-FC,tao2016siamese,SiamRPN,SiamRPN++,SiamRCNN,TransformerTrack,TransT,dong2018triplet,he2018twofold,zhu2018distractor,zhang2019deeper}, re-identification~\cite{chung2017two,zheng2019re,wu2018and,shen2017deep,Shen_2017_ICCV,TransReID}, and few-shot recognition~\cite{SiameseOneshotImageRecognition,ProtoNet,RelationNetwork,NEURIPS2019_92af93f7,MetaDETR}. Empirical results have shown that Siamese-based architectures, which project both inputs into the same embedding space using identical sub-networks, perform extraordinarily well on these matching tasks. This is because Siamese-based architectures can produce comparable feature vectors with aligned semantics for the inputs, thus facilitating the computation of similarities among them.

Our work is motivated by the success of Siamese-based architectures in various matching-involved tasks. We interpret DETR's cross-attention as a `matching and feature distillation' process and leverage the philosophy of Siamese networks to facilitate the matching procedure. We believe that it is essential to impose aligned semantics between object queries and encoded image features so that the similarity between them can be efficiently and accurately computed for the accelerated convergence of DETR.

\section{A Brief Review of DETR}   \label{sec:DETR_Review}

Since our proposed SAM-DETR++ is developed on top of DETR~\cite{DETR} for its accelerated convergence and superior detection performance, we first briefly review the basic architecture of DETR~\cite{DETR} before introducing our method.

Unlike ConvNet-based object detectors~\cite{FasterRCNN,SSD,YOLO9000,FCOS,zhou2019objects} that address object detection by solving surrogate classification and regression tasks, DETR~\cite{DETR} directly formulates object detection as a set prediction problem. The pipeline of DETR is simple: a backbone network, a Transformer encoder, and a Transformer decoder. Given an input image $\mathbf{I} \in \mathbb{R}^{H_{0} \times {W_{0}} \times 3}$, the backbone network and the Transformer encoder produce the encoded features for the input image $\mathbf{F} \in \mathbb{R}^{HW \times d}$, in which $d$ denotes the number of feature channels, and $H_{0}$, $W_{0}$ and $H$, $W$ denote the spatial sizes of the image and the encoded features, respectively. After that, the Transformer decoder takes the encoded image features $\mathbf{F}$ and a small set of object queries $\mathbf{Q} \in \mathbb{R}^{N \times d}$ as input, and then produces the detection results. Here, $N$ denotes the number of object queries, which is typically set to 100$\sim$300~\cite{DETR,DeformableDETR,ConditionalDETR,SMCA-DETR,Meta-DETR_firstversion,AnchorDETR,DABDETR,SAM-DETR,DA-DETR,CF-DETR,DynamicDETR,DINO}.

The Transformer decoder consists of multiple decoder layers, in which object queries are sequentially processed by a self-attention module, a cross-attention module, and a feed-forward network (FFN) to produce the outputs. The object queries output by each decoder layer are further fed into the subsequent layers and go through a Multi-Layer Perceptron (MLP) to produce detection predictions. The cross-attention module is the key element in the Transformer decoder, in which object queries interact with the encoded image features. As discussed in Section~\ref{sec:introduction} and illustrated in Fig.~\ref{fig:fig1}, the cross-attention module can be interpreted as a `matching and feature distillation' process: object queries first search for the relevant regions to match, then distill instance-level features from the matched regions to generate detection predictions. We formulate and interpret cross-attention as:
\begin{equation}  \label{eq:1}
\mathbf{Q^\prime} = \underbrace{\overbrace{{\rm Softmax}(\frac{(\mathbf{Q W_{\rm q}})(\mathbf{F W_{\rm k}})^{\rm T}}{\sqrt{d}})}^{\text{to match relevant regions}} (\mathbf{F W_{\rm v}})}_{\text{to distill features from relevant regions}},
\end{equation}
where $\mathbf{W_{\rm q}}$, $\mathbf{W_{\rm k}}$, and $\mathbf{W_{\rm v}}$ denote the linear projections for query, key, and value, respectively, in the Transformer attention mechanism, and $\mathbf{Q^\prime} \in \mathbb{R}^{N \times d}$ denotes the cross-attention's generated object queries.

Preferably, the cross-attention's output object queries $\mathbf{Q^\prime}$ should contain instance-level features distilled from the relevant regions, which are used to produce detection predictions. However, as discussed before and also verified in \cite{DeformableDETR,ConditionalDETR,SMCA-DETR,AnchorDETR,DABDETR,SAM-DETR}, the object queries are initially equally matched to all spatial locations in the encoded image features and are very challenging to learn to focus on specific regions properly. The matching difficulty with unaligned semantics is the key root for DETR's slow convergence.

\section{SAM-DETR++}   \label{sec:method}

This section presents the proposed \textit{SAM-DETR++} in detail, which greatly accelerates the convergence speed of the original DETR~\cite{DETR} with a plug-and-play module to achieve semantic-aligned matching. The proposed SAM-DETR++ can also effectively fuse multi-scale features in a coarse-to-fine manner on the basis of the introduced semantic-aligned matching mechanism. In addition, being like a plug and play, SAM-DETR++ can be integrated with multiple DETR's convergence solutions to achieve even faster convergence and superior detection accuracy.

\begin{figure*}[t!] 
\begin{center}
   \includegraphics[width=0.985\linewidth]{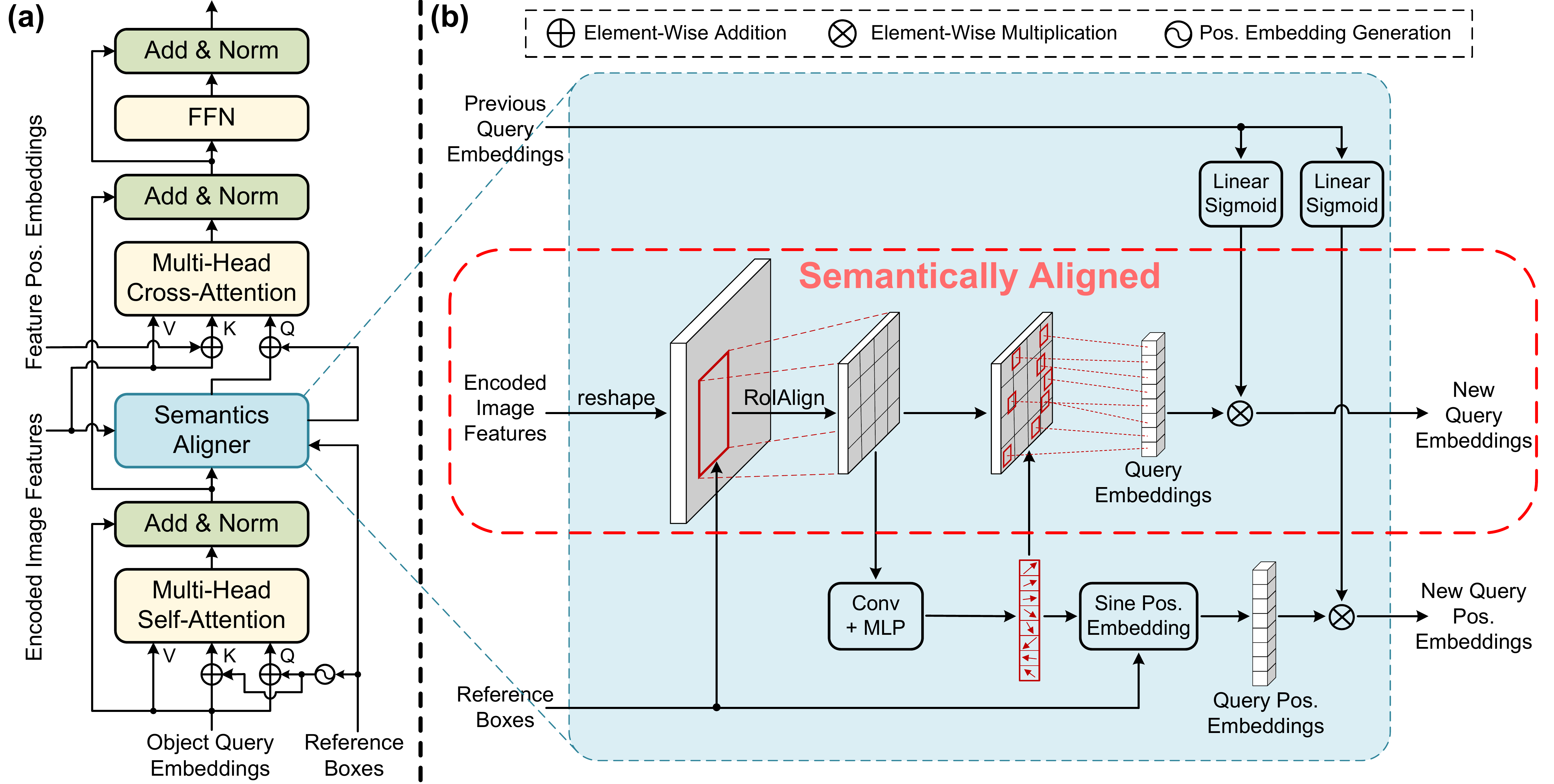}
\end{center}
\vspace{-5mm}
   \caption{
   \textbf{\textit{(a)} The overview of one Transformer decoder layer of the proposed \textit{SAM-DETR++}}. It models a learnable \textit{reference box} for each object query, whose center location is used to generate the corresponding positional embeddings. It also appends a \textit{Semantics Aligner} ahead of the cross-attention module, which generates new object queries that are semantically aligned with the encoded image features, thus facilitating their subsequent matching process within cross-attention. \textbf{\textit{\;(b)} The architecture of the proposed \textit{Semantics Aligner}.} Only one object query is presented for concise illustration. Semantics Aligner first extracts region features from the reference boxes' corresponding regions via RoIAlign, which are used to predict the coordinates of several representative keypoints with the most 
  distinctive features. The features from these representative keypoints are then sampled as the new query embeddings, which are semantically aligned with the encoded image features. Finally, the new query embeddings are further reweighted by the previous query embeddings to incorporate useful information from them.
   }
\label{fig:architecture}
\end{figure*}

\subsection{Overview}  \label{sec:method_overview}

Motivated by the observations discussed before, our proposed SAM-DETR++ aims to address DETR's slow convergence issue by relieving the complexity of the matching process as illustrated in Fig.\,\ref{fig:fig1}. The core idea is to project both object queries and encoded image features into the same feature embedding space, thus imposing a strong prior for each object query to focus on its relevant region with similar semantics within the cross-attention module. To achieve this, SAM-DETR++ only makes some minor modifications to the decoder of the original DETR~\cite{DETR}.

Fig.\,\ref{fig:architecture}\,(a) illustrates the overall architecture of the Transformer decoder of SAM-DETR++. Same as the original DETR~\cite{DETR}, each decoder layer is repeated six times, with zeros as input for the first layer and previous layer's outputs as input for the subsequent layers. As shown in Fig.\,\ref{fig:architecture}\,(a), in each decoder layer, a plug-and-play module, named \textit{Semantics Aligner}, is appended ahead of the cross-attention module for imposing aligned semantics between object queries and encoded image features. Besides, SAM-DETR++ also models a learnable reference box for each object query instead of directly modeling its query positional embeddings. The learnable reference boxes $\mathbf{R}_{\rm box} \in \mathbb{R}^{N \times 4}$ are modeled at the first decoder layer, representing object queries' initial locations. With the spatial guidance of these reference boxes, the proposed Semantics Aligner takes the previous object query embeddings $\mathbf{Q}$ and the encoded image features $\mathbf{F}$ as inputs to obtain new object query embeddings $\mathbf{Q^{\rm new}}$ and their corresponding positional embeddings $\mathbf{Q_{\rm pos}^{\rm new}}$, and then feed them into the subsequent cross-attention module. In this way, the generated embeddings $\mathbf{Q^{\rm new}}$ lie within the same embedding space as the encoded image features $\mathbf{F}$, which facilitates the following matching process between them, allowing object queries to quickly and properly attend to relevant regions with similar semantics in the encoded image features. It is worth noting that no modification is made to the other components in DETR's decoder layers, including multi-head self-attention, multi-head cross-attention, and FFN.

\subsection{Semantics Aligner}    \label{sec:SAM_DETR_semantic_aligner}

The detailed architecture of the appended \textit{Semantics Aligner} is illustrated in Fig.~\ref{fig:architecture}\,(b).

\vspace{+1.5mm}
\noindent
\textbf{Semantic-Aligned Matching.\;\;\;}
As formulated in Eq.\,\ref{eq:1} and illustrated in Fig.\,\ref{fig:fig1}\,(left), the cross-attention module uses dot-product to produce the attention heatmaps that represent the matching between object queries and encoded image features. It is natural and intuitive to adopt dot-product for generating the attention heatmaps, as dot-product is a good metric for the similarity between two feature vectors, which encourages object queries to have higher attention weights for regions with higher similarities. However, as illustrated in Fig.~\ref{fig:fig1}\,(right), the modules between cross-attentions project object queries into a different feature embedding space from that of the encoded image features, leading to unaligned semantics between them. The unaligned semantics causes each object query to almost equally match all spatial locations within the encoded image features at initialization, adding substantial complexity for learning a meaningful matching between them.

Motivated by these observations, we design Semantics Aligner to ensure that object query embeddings are within the same feature embedding space as encoded image features before being processed by cross-attention. This guarantees that the dot-product between query embeddings and encoded image features is always a meaningful measurement of similarity without the need to be explicitly learned, which imposes a prior for object queries to match relevant regions with similar semantics and reduces the matching difficulty.

The alignment of semantics is achieved by re-sampling new object query embeddings from the encoded image features. Concretely, as shown in Fig.\,\ref{fig:architecture}\,(b), Semantics Aligner first restores the encoded image features' spatial dimensions from 1D sequences $HW\,\times\,d$ to 2D feature maps $H\,\times W\,\times\,d$. Then, Semantics Aligner extracts region features $\mathbf{F_{\rm R}} \in \mathbb{R}^{N \times 7 \times 7 \times d}$ from the encoded image features via RoIAlign~\cite{MaskRCNN} from the corresponding reference boxes. Finally, the new object query embeddings $\mathbf{Q^{\rm new}}$ and new query positional embeddings $\mathbf{Q^{\rm new}_{\rm pos}}$ are obtained by re-sampling features from $\mathbf{F_{\rm R}}$. Mathematically, Semantics Aligner can be formulated at a high level as:
\begin{equation}
  \label{eq:2}
\mathbf{F_{\rm R}} = \text{RoIAlign}(\mathbf{F}, \mathbf{R}_{\rm box}) ,
\end{equation}
\begin{equation}
  \label{eq:3}
\mathbf{Q^{\rm new}}, \mathbf{Q^{\rm new}_{\rm pos}} = \text{Re-Sample}(\mathbf{F_{\rm R}}, \mathbf{R}_{\rm box}, \mathbf{Q}) .
\end{equation}
As the re-sampling procedure does not involve any projection (\textit{e.g.}, ConvNet or MLP), the new object query embeddings $\mathbf{Q^{\rm new}}$ always lie within the same feature embedding space as the encoded image features $\mathbf{F}$, which encourages object queries to focus on semantically similar regions in the following cross-attention module. The design choice for the re-sampling procedure is to be detailed later.

\begin{figure}[t!] 
\begin{center}
   \includegraphics[width=1.0\linewidth]{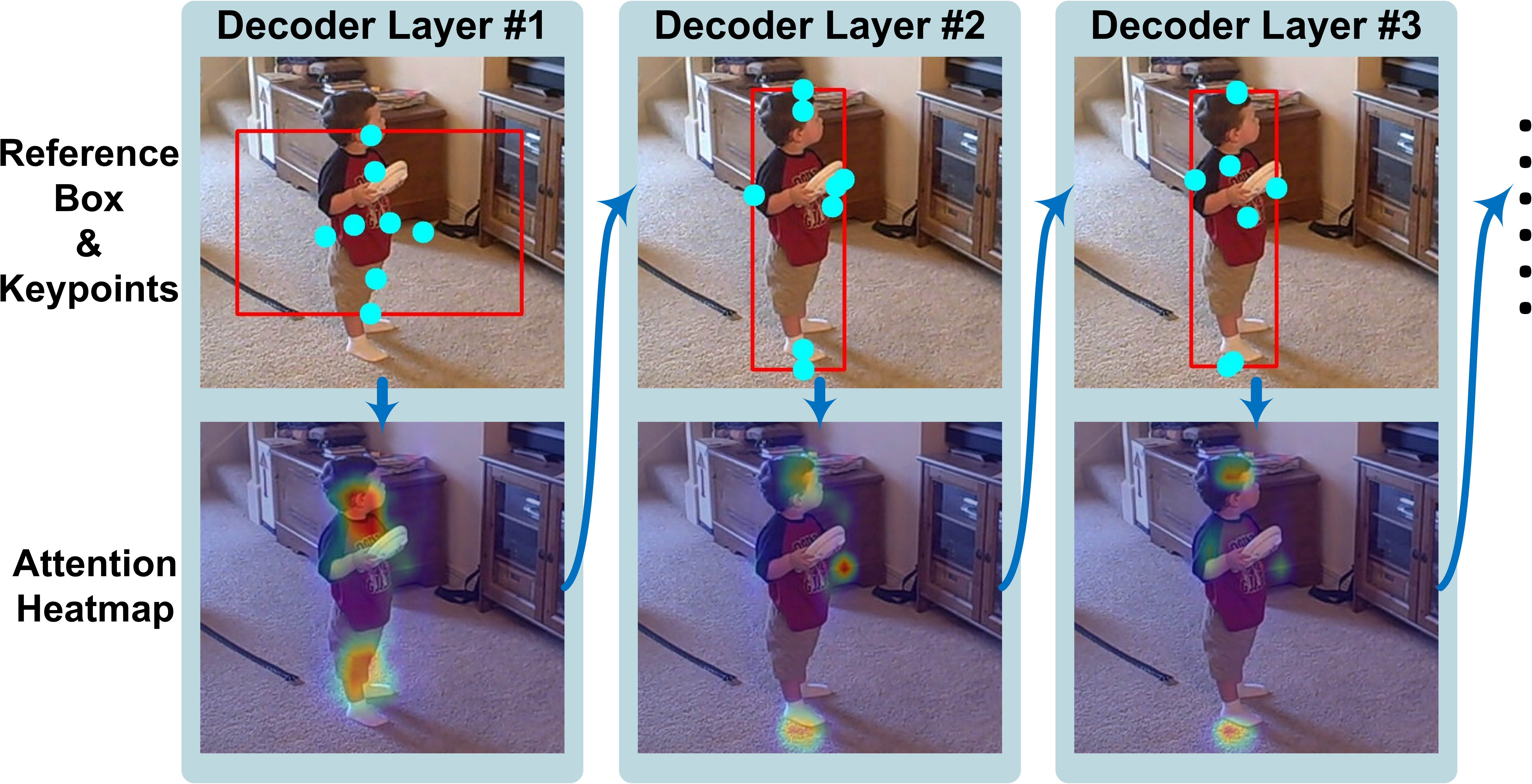}
\end{center}
\vspace{-2.5mm}
\caption{
Each decoder layer in SAM-DETR++ searches multiple representative keypoints (cyan dots) within each reference box (red box), and uses their features for semantic-aligned matching. As detection proceeds, the keypoints gradually fall on salient and semantically meaningful locations, and the attention heatmaps gradually become more precise.
}
\label{fig:keypoint_vis}
\end{figure}

\vspace{+2.5mm}
\noindent
\textbf{Semantic-Aligned Matching with Multiple Representative Keypoints.\;\;\;}
The re-sampling process can be easily accomplished by simple operations like applying global average-pooling or global max-pooling on region features $\mathbf{F_{\rm R}}$. But instead, we propose a more sophisticated approach to re-sample new object query embeddings, inspired by the prior works~\cite{ExtremeNet,reppoints,wu2020cascade,reppointsv2,DETR,DefectGAN,ConditionalDETR} that identify the importance of objects' representative keypoints in object detection. Specifically, Semantics Aligner explicitly searches for multiple representative keypoints for each object query and extracts their features for the aforementioned semantic-aligned matching. Such design can naturally fit in the multi-head attention mechanism~\cite{transformer} without any modification, enabling every attention head to produce different weights to focus on different parts.

Here, we denote the number of attention heads as $M$, which is set to 8 in most DETR-based object detectors~\cite{DETR,DeformableDETR,ConditionalDETR,SMCA-DETR,Meta-DETR,AnchorDETR,DABDETR,SAM-DETR}. $M$ is also the number of representative keypoints to search for each object query. As shown in Fig.\,\ref{fig:architecture}\,(b), after retrieving region features $\mathbf{F_{\rm R}}$, Semantics Aligner adopts a ConvNet followed by an MLP to predict the spatial locations of the $M$ keypoints for each object query, representing the locations that are crucial for recognizing and localizing the potential objects, which can be formulated as:
\begin{equation}
\mathbf{R}_{\rm SP} = \text{MLP}(\text{ConvNet}( \mathbf{F_{\rm R}})),
\end{equation}
where $\mathbf{R}_{\rm SP} \in \mathbb{R}^{N \times M \times 2}$ denotes the coordinates of $M$ keypoints for each of the $N$ object queries.
Note that the predicted coordinates are constrained to be inside their corresponding reference boxes. With the predicted $\mathbf{R}_{\rm SP}$, the features of these representative keypoints can be then sampled from $\mathbf{F_{\rm R}}$ via bi-linear interpolation. Semantics Aligner finally concatenates the $M$ sampled features vectors corresponding to the $M$ representative keypoints as the new query embeddings, which are fed into the subsequent multi-head cross-attention so that each attention head can focus on features of one representative keypoint. Similarly, the object queries' corresponding positional embeddings can be computed using sinusoidal functions based on the keypoints' image-scale coordinates, and are then also concatenated to feed into the multi-head cross-attention module.
\begin{equation}
\mathbf{Q}^{\rm new \prime} = \text{Concat}(\{\mathbf{F_{\rm R}}[...,x,y,...] \text{ for } x,y \in \mathbf{R}_{\rm SP}\})
\end{equation}
\begin{equation}
\mathbf{Q}^{\rm new \prime}_{\rm pos} = \text{Concat}(\text{Sinusoidal}{(\mathbf{R}_{\rm box}, \mathbf{R}_{\rm SP})})
\end{equation}

Fig.\,\ref{fig:keypoint_vis} visualizes the searched representative keypoints and the attention heatmaps produced by cross-attention. It can be observed that the search keypoints gradually fall on the salient positions with rich semantics (\textit{e.g.}, head and extremities) as detection proceeds. In addition, the attention heatmaps also gradually become more precise and focus on those semantically meaningful regions. These results validate the effectiveness of searching and exploiting keypoint features for semantic-aligned matching in enhancing its representation capacity.

\vspace{+3.0mm}
\noindent
\textbf{Feature Reweighting with Previous Query Embeddings.\;\;\;}
So far, Semantics Aligner can generate new object query embeddings $\mathbf{Q}^{\rm new \prime}$ with aligned semantics with the encoded image features. However, it also brings one issue: the cross-attention cannot leverage the previous query embeddings $\mathbf{Q}$ that contain valuable information for detection. To mitigate this issue, Semantics Aligner further receives the previous query emebddings $\mathbf{Q}$ as inputs to produce a set of reweighting coefficients via `Linear Projection + Sigmoid'. The reweighting coefficients are applied to new query embeddings $\mathbf{Q}^{\rm new \prime}$ and their positional embeddings $\mathbf{Q}^{\rm new \prime}_{\rm pos}$ through element-wise multiplication, highlighting those important features. As a result, the useful information from previous query embeddings can be effectively leveraged in cross-attention with the introduced semantic-aligned matching mechanism. Note that feature reweighting does not affect the aligned semantics of query embeddings, as it does not perform any projection on the query embeddings. The described feature reweighting process can be formulated as:
\begin{align}
\mathbf{Q}^{\rm new} = \mathbf{Q}^{\rm new \prime} \otimes \sigma (\mathbf{Q} \mathbf{W_{\rm RW1}}),  \\
\mathbf{Q}^{\rm new}_{\rm pos} = \mathbf{Q}^{\rm new \prime}_{\rm pos} \otimes \sigma (\mathbf{Q} \mathbf{W_{\rm RW2}}),
\end{align}
where $\mathbf{W_{\rm RW1}}$ and $\mathbf{W_{\rm RW2}}$ are the learnable parameters for linear projections, $\sigma(\cdot)$ denotes sigmoid function, and \,$\otimes$ denotes element-wise multiplication.

\begin{figure}[t!] 
\begin{center}
   \includegraphics[width=0.88\linewidth]{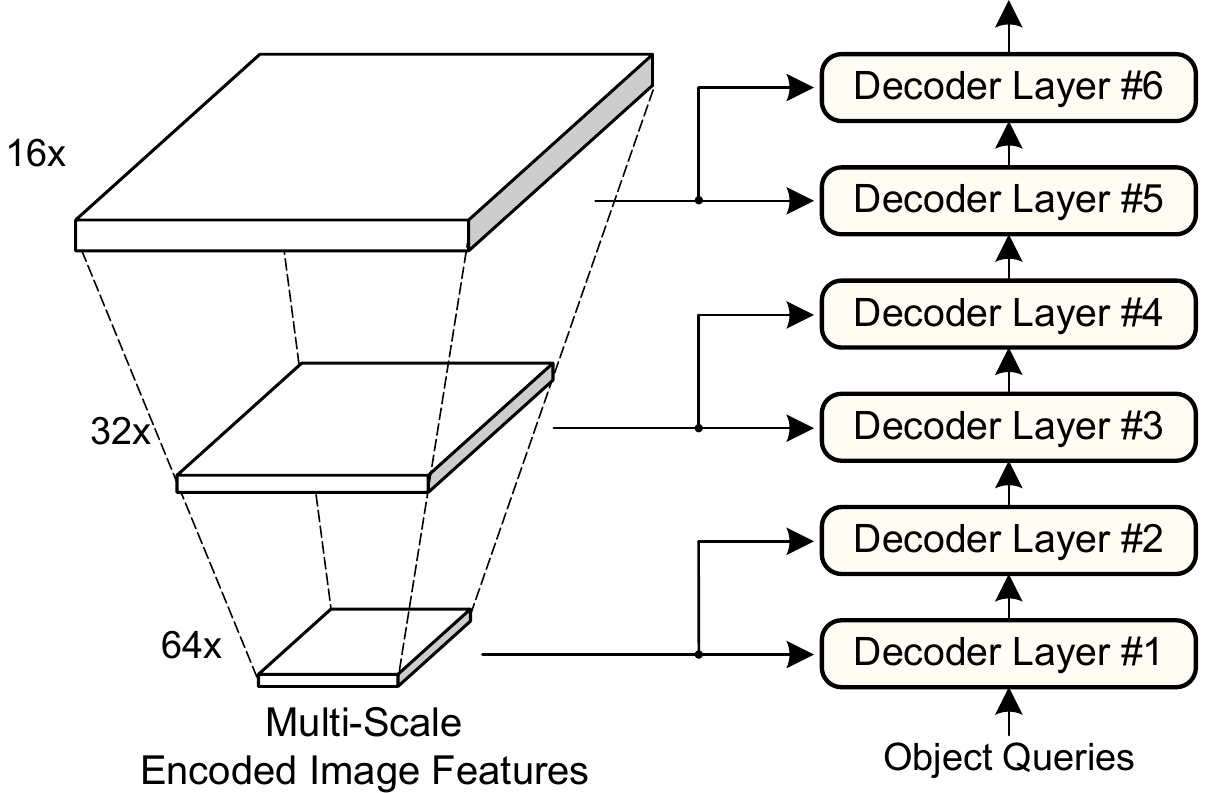}
\end{center}
\vspace{-2.5mm}
\caption{
The proposed SAM-DETR++ can be extended to fuse multi-scale features by simply feeding different feature scales into different decoder layers in a coarse-to-fine manner. Thanks to the introduced semantic-aligned matching mechanism, SAM-DETR++ can effectively fuse multi-scale features that are inherently unaligned in feature semantics.
}
\label{fig:multiscale}
\end{figure}

\subsection{Multi-Scale Feature Fusion with Aligned Semantics}

Detecting objects of vastly different scales has always been one major challenge in object detection. Modern ConvNet-based object detectors (\textit{e.g.,} Faster R-CNN\cite{FasterRCNN}\;w/\;FPN\cite{FPN}, M2Det\cite{m2det}, EfficientDet\cite{efficientdet}) usually incorporate multi-scale features to accommodate this issue by representing objects at different scales in a `divide and conquer' manner, which reduces the representation complexity and achieves superior detection accuracy and faster convergence. With this motivation, we further extend the proposed semantic-aligned matching strategy to fuse multi-scale features in a coarse-to-fine manner.

As shown in Fig.\,\ref{fig:multiscale}, the proposed method to fuse multi-scale features is simple and concise. Considering the cascade nature of DETR's decoder, we feed the features of different scales into different stages of the decoder, making the detection pipeline a coarse-to-fine refinement process. Concretely, the first two decoder layers receive the coarsest feature maps to reduce search space for initial localization; the subsequent two layers receive finer feature maps for more precise localization; the last two layers receive high-resolution feature maps for detecting tiny objects. This simple and effective design does not introduce extra parameters but allows SAM-DETR++ to adaptively fuse multi-scale features to represent objects at different scales, thus greatly lowering the learning complexity.

It is worth noting that it is our proposed semantic-aligned matching mechanism that enables this simple approach to fuse multi-scale features effectively. Experiments in Section~\ref{sec:ablation_study} show that without the proposed semantic-aligned matching, directly fusing multi-scale features does not bring a clear performance gain. This is because of the inevitable unaligned semantics across different feature scales, which causes extra complexity in the matching processes between object queries and encoded image features, as discussed before. Our introduced Semantics Aligner alleviates this issue by explicitly aligning the semantics between object query embeddings and encoded image features at all decoder layers, thus enabling the effective fuse of features from different scales.

\subsection{Removing Dropout in Transformer} \label{sec:SAMDETR_removedropout}

Most existing DETR-like detectors~\cite{DETR,DeformableDETR,Meta-DETR,SMCA-DETR,ConditionalDETR,SparseDETR}, including the \textit{CVPR'\,2022} conference version of SAM-DETR~\cite{SAM-DETR}, contain dropout~\cite{srivastava2014dropout} in the Transformer encoder-decoder architecture~\cite{transformer}. Dropout~\cite{srivastava2014dropout} is included to mitigate the overfitting issue in natural language processing (NLP) tasks. However, in the task of object detection within images, we empirically find that dropout does not mitigate overfitting but harms the performance of object detection. We conjecture that this is largely attributed to the unique property of object detection within 2D images, where adjacent pixels within feature maps are strongly correlated. Therefore, we incorporate a minor tweak to remove the dropout in the Transformer, which yields better performance at no computational cost. The effectiveness of this modification is verified in Section~\ref{sec:ablation_study}.

\begin{figure*}[t] 
\begin{center}
   \includegraphics[width=1.0\linewidth]{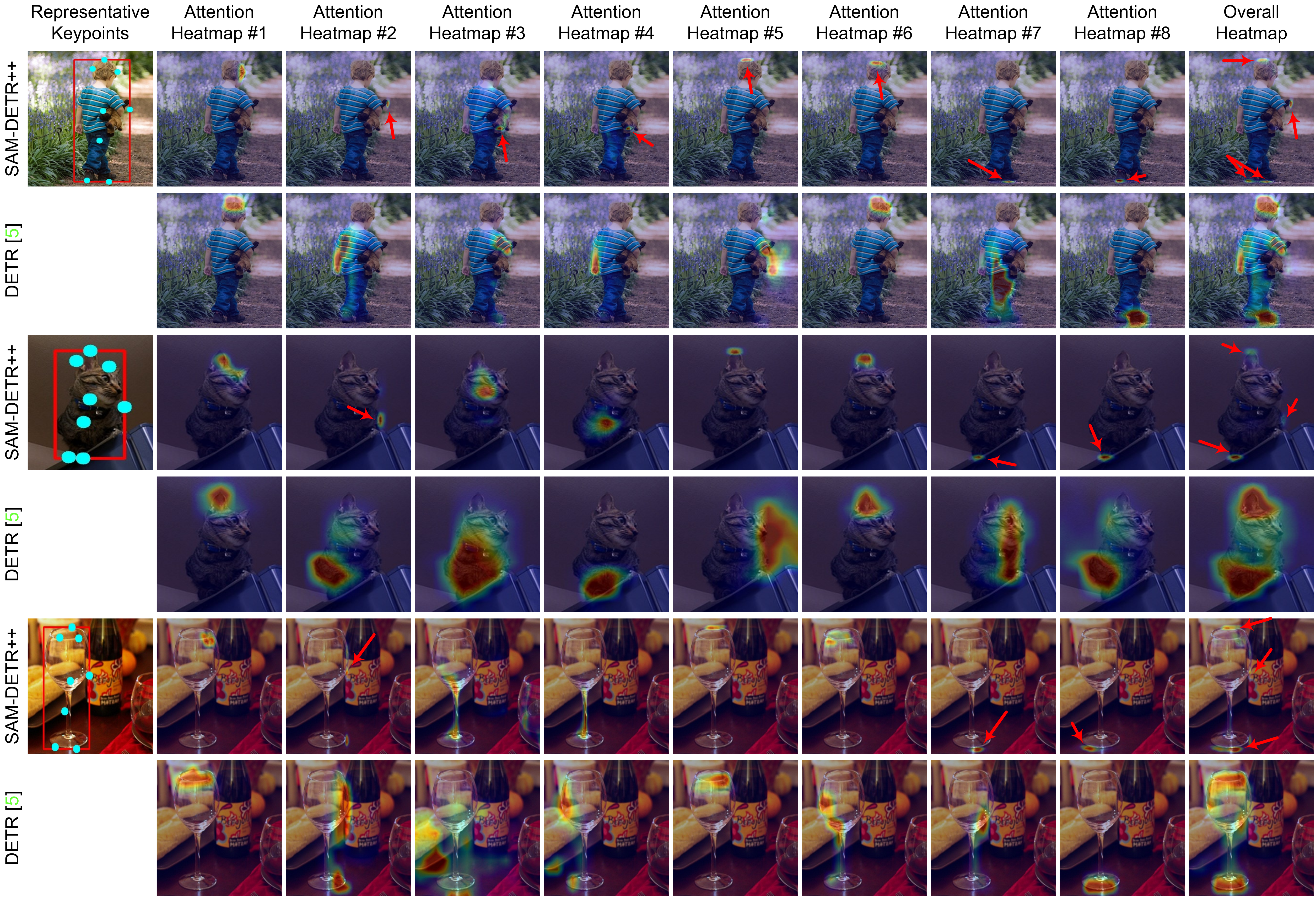}
\end{center}
\vspace{-5.5mm}
   \caption{
   Visualization of the searched representative keypoints and the attention heatmaps of different attention heads in cross-attention from our proposed SAM-DETR++. The searched representative keypoints mostly fall around objects of interest, and typically fall on the positions with the most distinctive features for recognition or localization, like object extremities or central points. Our method's attention heatmaps are much more focused compared with the original DETR without semantic-aligned matching, which proves the effectiveness of our approach in relieving the complication in the matching processes between object queries and encoded image features, which accelerates DETR's convergence. Red-colored arrows highlight those fine details in attention heatmaps. Zoom-in may be required to view details.
    }
\label{fig:visualization}
\vspace{-1mm}
\end{figure*}

\subsection{Compatibility with Existing Convergence Solutions}

As illustrated in Fig.\,\ref{fig:architecture}\,(a), SAM-DETR++ only appends a plug-and-play module into the Transformer decoder layer, leaving most other operations unchanged. Besides, SAM-DETR++ speeds up DETR's training convergence from a distinct perspective from existing convergence solutions. These properties make it easy and effective to integrate SAM-DETR++ with other approaches to achieve even faster convergence and superior detection accuracy. Here, we integrate our method with two recent works to validate the strong compatibility of SAM-DETR++.

\subsubsection{Compatibility with SMCA-DETR}

SMCA-DETR~\cite{SMCA-DETR} replaces DETR's original cross-attention module with Spatially Modulated Co-Attention (SMCA), which estimates the position of each object query, and then applies a series of 2D-Gaussian weight maps to constrain the attention responses in different attention heads. Both the center locations and the scales for SMCA's 2D-Gaussian weight maps are predicted from the corresponding object query embeddings. SMCA~\cite{SMCA-DETR} effectively accelerates DETR's convergence by imposing spatial constraints to the SMCA module.

To integrate SMCA~\cite{SMCA-DETR} into our proposed SAM-DETR++, we make one minor modification to the SMCA mechanism: we adopt the coordinates of the \textit{M} representative keypoints as the central locations for the 2D Gaussian weight maps. The scales of the weight maps are also predicted from the region features in parallel to the central locations. Experiment results in Section~\ref{sec:experiment_quantitative} validate the complementary effect between our proposed SAM-DETR++ and SMCA~\cite{SMCA-DETR}.

\subsubsection{Compatibility with DN-DETR}

The recently proposed DN-DETR~\cite{DN-DETR} introduces a novel de-noising training strategy to speed up DETR's training procedure, which is also complementary to our approach without any adaptation. Experiment results in Section~\ref{sec:experiment_quantitative} also validate the complementary effect between our proposed SAM-DETR++ and DN-DETR~\cite{DN-DETR}.

\begin{table*}[t]
\begin{center}
\centering
\caption{Object detection performance under the 12-epoch (1x) training schedule on COCO\;val\,2017.}
\label{tab:exp_1x}
\vspace{-4.0mm}
\setlength{\tabcolsep}{4.86pt}
\resizebox{1.0\textwidth}{!}{
\begin{tabular}[t]{l|c|ccc|cccccc}
\toprule[1.35pt]
Method & multi-scale & \#Epochs & \#Params\,(M) & GFLOPs & AP & AP$_{\rm 50}$ & AP$_{\rm 75}$ & AP$_{\rm S}$ & AP$_{\rm M}$ & AP$_{\rm L}$ \\

\midrule[1.0pt]

\rowcolor{Gray} \multicolumn{11}{l}{\textbf{\textit{$\bullet$ Backbone: ResNet-50\; (Single-Scale Features)}}} \\

Faster-R-CNN-R50~\cite{FasterRCNN} & & 12 & 34 & 547 & 35.7 & 56.1 & 38.0 & \textbf{19.2} & \textbf{40.9} & 48.7 \\

DETR-R50~\cite{DETR}\;$\ddag$ &  & 12 & 41 & 86 & 22.3 & 39.5 & 22.2 & 6.6 & 22.8 & 36.6 \\

Deformable-DETR-R50~\cite{DeformableDETR} & & 12 & 34 & 78 & 31.8 & 51.4 & 33.5 & 15.0 & 35.7 & 44.7 \\

Conditional-DETR-R50~\cite{ConditionalDETR} & & 12 & 44 & 90 & 32.2 & 52.1 & 33.4 & 13.9 & 34.5 & 48.7 \\

SMCA-DETR-R50~\cite{SMCA-DETR} & & 12 & 42 & 86 & 31.6 & 51.7 & 33.1 & 14.1 & 34.4 & 46.5 \\


\textbf{SAM-DETR-R50 (Ours)} & & 12 & 57 & 107 & 34.2 & 55.8 & 35.3 & 15.0 & 37.7 & 52.5 \\

\textbf{SAM-DETR-R50\;w/\;SMCA (Ours)} & & 12 & 57 & 107 & \textbf{37.0} & \textbf{58.0} & \textbf{38.5} & 17.8 & 40.3 & \textbf{56.1} \\

\midrule[0.68pt]

\rowcolor{Gray} \multicolumn{11}{l}{\textbf{\textit{$\bullet$ Backbone: ResNet-50-DC5\; (High-Resolution Features)}}} \\

Faster-R-CNN-R50-DC5~\cite{FasterRCNN} & & 12 & 166 & 320 & 37.3 & 58.8 & 39.7 & 20.1 & 41.7 & 50.0 \\

DETR-R50-DC5~\cite{DETR}\;$\ddag$ & & 12 & 41 & 187 & 25.9 & 44.4 & 26.0 & 7.9 & 27.1 & 41.4 \\

Deformable-DETR-R50-DC5~\cite{DeformableDETR} & & 12 & 34 & 128 & 34.9 & 54.3 & 37.6 & 19.0 & 38.9 & 47.5 \\

Conditional-DETR-R50-DC5~\cite{ConditionalDETR} & & 12 & 44 & 195 & 35.9 & 55.8 & 38.2 & 17.8 & 38.8 & 52.0 \\

SMCA-DETR-R50-DC5~\cite{SMCA-DETR} & & 12 & 42 & 187 & 32.5 & 52.8 & 33.9 & 14.2 & 35.4 & 48.1 \\

Anchor-DETR-R50-DC5~\cite{AnchorDETR} & & 12 & 39 & 151 & 37.1 & 57.8 & 39.1 & 19.0 & 40.8 & 51.4 \\

DAB-DETR-R50-DC5~\cite{DABDETR} & & 12 & 44 & 216 & 38.0 & 60.3 & 39.8 & 19.2 & 40.9 & 55.4  \\ 

DN-DETR-R50-DC5~\cite{DN-DETR} & & 12 & 44 & 216 & 41.7 & 61.4 & 44.1 & 21.2 & 45.0 & 60.2  \\ 

\textbf{SAM-DETR-R50-DC5 (Ours)} & & 12 & 57 & 229 & 39.1 & 59.9 & 41.2 & 20.9 & 42.8 & 55.5 \\

\textbf{SAM-DETR-R50-DC5\;w/\;SMCA (Ours)} & & 12 & 57 & 229 & 41.3 & 61.6 & 43.6 & 22.1 & 44.9 & 59.2 \\

\textbf{SAM-DETR-R50-DC5\;w/\;DN (Ours)} & & 12 & 57 & 229 & 42.3 & 61.7 & 45.2 & 22.8 & 45.7 & 60.0 \\

\textbf{SAM-DETR-R50-DC5\;w/\;SMCA\,+\,DN (Ours)} & & 12 & 57 & 229 & \textbf{43.7} & \textbf{63.0} & \textbf{46.8} & \textbf{24.3} & \textbf{47.4} & \textbf{61.4} \\

\midrule[0.68pt]

\rowcolor{Gray} \multicolumn{11}{l}{\textbf{\textit{$\bullet$ Backbone: ResNet-50\; (Multi-Scale Features)}}} \\

Faster-R-CNN-R50-FPN~\cite{FasterRCNN,FPN} & $\checkmark$ & 12 & 42 & 180 & 37.9 & 58.8 & 41.1 & 22.4 & 41.1 & 49.1 \\

Cascade-R-CNN-R50-FPN~\cite{CascadeRCNN,FPN} & $\checkmark$ & 12 & 69 & 230 & 40.4 & 58.9 & 44.1 & 22.8 & 43.7 & 54.0 \\

FCOS-R50~\cite{FCOS} & $\checkmark$ & 12 & 32 & 201 & 38.6 & 57.2 & 41.7 & 23.5 & 42.8 & 48.9 \\

Sparse-R-CNN-R50-FPN~\cite{SparseRCNN} & $\checkmark$ & 12 & 106 & 166 & 40.1 & 59.4 & 43.5 & 22.9 & 43.6 & 52.9 \\

Deformable-DETR-R50~\cite{DeformableDETR} & $\checkmark$ & 12 & 40 & 173 & 37.2 & 55.5 & 40.5 & 21.1 & 40.7 & 50.5 \\

SMCA-DETR-R50~\cite{SMCA-DETR} & $\checkmark$ & 12 & 40 & 152 & 35.0 & 54.1 & 37.8 & 18.7 & 37.7 & 48.1 \\

\textbf{SAM-DETR++-R50\;(Ours)} & $\checkmark$ & 12 & 55 & 203 & 41.9 & 60.5 & 45.3 & 24.6 & 45.5 & 57.4 \\

\textbf{SAM-DETR++-R50\;w/\;SMCA\;(Ours)} & $\checkmark$ & 12 & 55 & 203 & 43.2 & 61.5 & 46.5 & 25.5 & 46.5 & 58.6 \\

\textbf{SAM-DETR++-R50\;w/\;SMCA\,+\,DN\;(Ours)} & $\checkmark$ & 12 & 55 & 203 & \textbf{44.8} & \textbf{62.6} & \textbf{47.9} & \textbf{26.7} & \textbf{48.2} & \textbf{60.9} \\

\bottomrule[1.35pt]
\end{tabular}}
\vspace{+0.25mm}
  \begin{tablenotes}
    \item[1] \;\;\;`$\ddag$' denotes the original DETR baseline~\cite{DETR} with increased number of object query (100$\rightarrow$300) and focal loss as the classification loss function.
  \end{tablenotes}
\end{center}
\end{table*}

\section{Experiments}   \label{sec:experiments}

\subsection{Experiment Setup}

\vspace{+0.5mm}
\noindent
\textbf{Dataset and Evaluation Metrics.\;\;}
Following prior works~\cite{DETR,DeformableDETR,ConditionalDETR,SMCA-DETR,AnchorDETR,SAM-DETR,DABDETR}, we mainly perform the experiments on the COCO 2017 dataset~\cite{MSCOCO}, using the $\sim$117k images in the train2017 set for training and the 5k images in the val2017 set for evaluation. We adopt the standard metrics defined by COCO to evaluate the performance of object detection.

\vspace{+1.5mm}
\noindent
\textbf{Implementation Details.\;\;\;}
SAM-DETR++'s implementation details mostly align with the original DETR~\cite{DETR} and other prior works~\cite{DeformableDETR,ConditionalDETR,SMCA-DETR,AnchorDETR,DABDETR}. We use ImageNet-pretrained~\cite{imagenet} ResNet-50~\cite{resnet} as the backbone network. All experiments are performed on servers with 8\,$\rm\times$\,Nvidia V100 GPUs. We train our models with AdamW optimizer~\cite{Adam,AdamW}. The batch size is set to 16 for training, except when ResNet-50-DC5 is used as the backbone, the batch size is set to 8. The initial learning rate is $\rm 1\!\times\!10^{-5}$ for the backbone parameters and $\rm 1\!\times\!10^{-4}$ for the other parameters. The weight decay is set to $\rm 1\!\times\!10^{-4}$. Two training schedules are experimented: \textit{(i)} the 12-epoch (1x) schedule that is widely adopted in ConvNet-based detectors~\cite{FasterRCNN,FPN,focalloss,FCOS}, where the learning rate decays at the 10$^{\text{th}}$ epoch; \textit{(ii)} the 50-epoch schedule that is often used in Transformer-based detectors~\cite{DeformableDETR,ConditionalDETR,SMCA-DETR,AnchorDETR,DABDETR}, where the learning rate decays at the 40$^{\text{th}}$ epoch. Model-related hyper-parameters (\textit{e.g.}, feature channel dimension, number of encoder and decoder layers) remain the same with DETR~\cite{DETR}, except we make two minor modifications following some recent works~\cite{DeformableDETR,Meta-DETR,ConditionalDETR,SMCA-DETR,AnchorDETR} to improve DETR's convergence speed: the number of object queries $N$ is increased from 100 to 300; the sigmoid focal loss~\cite{focalloss} is adopted as the classification loss instead of the cross-entropy loss. These two modifications are also applied to the original DETR~\cite{DETR} for a fair comparison with the baseline.

The same data augmentation used in prior works \cite{DETR,DeformableDETR,ConditionalDETR,AnchorDETR,DABDETR,SAM-DETR} is adopted, which includes random resize, horizontal flip, and random crop. We constrain the training images' longest sides to be less or equal than 1333 pixels and the shortest sides to be larger or equal than 480 pixels.

\subsection{Visualization and Analysis}   \label{sec:experiment_vis}

Fig.\;\ref{fig:visualization} visualizes the representative keypoints searched by the proposed Semantics Aligner as well as their corresponding attention heatmaps generated from the subsequent multi-head cross-attention module. We also compare the attention heatmaps with the ones generated from the original DETR~\cite{DETR}. Results are obtained under the 12-epoch (1x) training schedule using ResNet-50~\cite{resnet}.

The visualization shows that the searched representative keypoints mostly fall around the target objects, and typically at those representative positions with the most distinctive features, such as object extremities or central points. The attention response heatmaps generated by the subsequent cross-attention modules also show high responses on those searched representative keypoints accordingly. In addition, compared with the original DETR~\cite{DETR}, our method shows clearly more precise and focused responses, which validates that the proposed semantic-aligned matching mechanism successfully facilitates the matching of object queries with appropriate regions for distillation of relevant instance-level features, thus accelerating DETR's convergence.

\begin{table*}[t!]
\begin{center}
\centering
\caption{Comparison with state-of-the-art object detectors on COCO\;val\,2017 under longer training schedules.}
\label{tab:exp_long}
\vspace{-4.486mm}
\setlength{\tabcolsep}{7.777pt}
\resizebox{0.995\textwidth}{!}{
\begin{tabular}[t]{l|ccc|cccccc}

\toprule[1.35pt]

Method & \#Epochs & \#Params\,(M) & GFLOPs & AP & AP$_{\rm 50}$ & AP$_{\rm 75}$ & AP$_{\rm S}$ & AP$_{\rm M}$ & AP$_{\rm L}$ \\

\midrule[1.0pt]

\rowcolor{Gray} \multicolumn{10}{l}{\textbf{\textit{$\bullet$ \;Baseline methods trained for extra-long epochs:}}} \\

Faster-R-CNN-R50-FPN~\cite{FasterRCNN,FPN} & 108 & 42 & 180 & 42.0 & 62.1 & 45.5 & 26.6 & 45.4 & 53.4 \\

DETR-R50~\cite{DETR} & 500 & 41 & 86 & 42.0 & 62.4 & 44.2 & 20.5 & 45.8 & 61.1 \\

DETR-R50-DC5~\cite{DETR} & 500 & 41 & 187 & 43.3 & 63.1 & 45.9 & 22.5 & 47.3 & 61.1 \\

\midrule[1.0pt]

\rowcolor{Gray} \multicolumn{10}{l}{\textbf{\textit{$\bullet$ \;ConvNet-based object detectors:}}} \\

Cascade-Mask-R-CNN-R50-FPN~\cite{CascadeRCNN} & 36 & 77 & 394 & 44.3 & 62.2 & 48.0 & 26.6 & 47.7 & 57.7 \\

TSP-FCOS-R50-FPN~\cite{TSPRCNN} & 36 & 52 & 189 & 43.1 & 62.3 & 47.0 & 26.6 & 46.8 & 55.9 \\

TSP-R-CNN-R50-FPN~\cite{TSPRCNN} & 36 & 64 & 188 & 43.8 & 63.3 & 48.3 & 28.6 & 46.9 & 55.7 \\

TSP-R-CNN-R50-FPN~\cite{TSPRCNN} & 96 & 64 & 188 & 45.0 & 64.5 & 49.6 & 29.7 & 47.7 & 58.0 \\

Sparse-R-CNN-R50-FPN~\cite{SparseRCNN} & 36 & 106 & 166 & 45.0 & 63.4 & 48.2 & 26.9 & 47.2 & 59.5 \\

\midrule[1.0pt]

\rowcolor{Gray} \multicolumn{10}{l}{\textbf{\textit{$\bullet$ \;Transformer-based object detectors:}}} \\

DETR-R50~\cite{DETR} $\ddag$ & 50 & 41 & 86 & 34.9 & 55.5 & 36.0 & 14.4 & 37.2 & 54.5 \\

DETR-R50-DC5~\cite{DETR} $\ddag$ & 50 & 41 & 187 & 36.7 & 57.6 & 38.2 & 15.4 & 39.8 & 56.3 \\

UP-DETR-R50~\cite{up-detr} & 150 & 41 & 86 & 40.5 & 60.8 & 42.6 & 19.0 & 44.4 & 60.0 \\

UP-DETR-R50~\cite{up-detr} & 300 & 41 & 86 & 42.8 & 63.0 & 45.3 & 20.8 & 47.1 & 61.7 \\

Deformable-DETR-R50~\cite{DeformableDETR} & 50 & 40 & 173 & 43.8 & 62.6 & 47.7 & 26.4 & 47.1 & 58.0 \\

Deformable-DETR-R50\;(two-stage)~\cite{DeformableDETR} & 50 & 40 & 173 & 46.2 & 65.2 & 50.0 & 28.8 & 49.2 & 61.7 \\

SMCA-DETR-R50~\cite{SMCA-DETR} & 50 & 40 & 152 & 43.7 & 63.6 & 47.2 & 24.2 & 47.0 & 60.4 \\

SMCA-DETR-R50~\cite{SMCA-DETR} & 108 & 40 & 152 & 45.6 & 65.5 & 49.1 & 25.9 & 49.3 & 62.6 \\

Conditional-DETR-R50~\cite{ConditionalDETR} & 50 & 44 & 90 & 40.9 & 61.8 & 43.3 & 20.8 & 44.6 & 59.2 \\

Conditional-DETR-R50~\cite{ConditionalDETR} & 108 & 44 & 90 & 43.0 & 64.0 & 45.7 & 22.7 & 46.7 & 61.5 \\

Conditional-DETR-R50-DC5~\cite{ConditionalDETR} & 50 & 44 & 195 & 43.8 & 64.4 & 46.7 & 24.0 & 47.6 & 60.7 \\

Conditional-DETR-R50-DC5~\cite{ConditionalDETR} & 108 & 44 & 195 & 45.1 & 65.4 & 48.5 & 25.3 & 49.0 & 62.2 \\

Anchor-DETR-R50~\cite{AnchorDETR} & 50 & 37 & 93 & 42.1 & 63.1 & 44.9 & 22.3 & 46.2 & 60.0 \\

Anchor-DETR-R50-DC5~\cite{AnchorDETR} & 50 & 37 & 172 & 44.2 & 64.7 & 47.5 & 24.7 & 48.2 & 60.6 \\

DAB-DETR-R50~\cite{DABDETR} & 50 & 44 & 94 & 42.2 & 63.1 & 44.7 & 21.5 & 45.7 & 60.3 \\

DAB-DETR-R50-DC5~\cite{DABDETR} & 50 & 44 & 202 & 44.5 & 65.1 & 47.7 & 25.3 & 48.2 & 62.3 \\

DAB-DETR-R50-DC5 (3\,Patterns)~\cite{DABDETR} & 50 & 44 & 216 & 45.7 & 66.2 & 49.0 & 26.1 & 49.4 & 63.1 \\

Sparse-DETR-R50~\cite{SparseDETR} & 50 & 41 & 136 & 46.3 & 66.0 & 50.1 & 29.0 & 49.5 & 60.8  \\

DN-DETR-R50~\cite{DN-DETR} & 50 & 44 & 94 & 44.1 & 64.4 & 46.7 & 22.9 & 48.0 & 63.4 \\ 

DN-DETR-R50-DC5~\cite{DN-DETR} & 50 & 44 & 202 & 46.3 & 66.4 & 49.7 & 26.7 & 50.0 & 64.3 \\ 

\textbf{SAM-DETR++-R50 (Ours)} & 50 & 55 & 203 & 47.5 & 66.5 & 51.3 & 29.3 & 50.8 & 62.7 \\

\textbf{SAM-DETR++-R50 w/\;SMCA (Ours)} & 50 & 55 & 203 & 48.0 & 66.6 & 52.2 & 29.9 & 51.5 & 64.6 \\

\textbf{SAM-DETR++-R50 w/\;SMCA\,+\,DN (Ours)} & 50 & 55 & 203 & \textbf{49.1} & \textbf{67.2} & \textbf{53.2} & \textbf{30.5} & \textbf{52.6} & \textbf{64.7} \\

\bottomrule[1.35pt]

\end{tabular}}
\vspace{+0.25mm}
  \begin{tablenotes}
    \item[1] \;\;\;`$\ddag$' denotes the original DETR baseline~\cite{DETR} with increased number of object query (100$\rightarrow$300) and focal loss as the classification loss function.
  \end{tablenotes}
\end{center}
\vspace{+0.88mm}
\end{table*}

\subsection{Experiment Results}   \label{sec:experiment_quantitative}

This subsection presents experiment results under two different training schedules. Here, we denote our proposed method with multi-scale feature fusion included as \verb|SAM-DETR++|, and denote our method without multi-scale feature fusion as \verb|SAM-DETR|. It is noteworthy that \verb|SAM-DETR| is identical to its \textit{CVPR'\,2022} conference version~\cite{SAM-DETR}, except that we remove dropout~\cite{srivastava2014dropout} as discussed in Section~\ref{sec:SAMDETR_removedropout}.

\vspace{+2mm}
\noindent
\textbf{Results under the 12-epoch (1x) Schedule.\;\;\;}
We first present object detection results under the short 12-epoch (1x) training schedule, which is widely used in conventional ConvNet-based object detectors~\cite{FasterRCNN,YOLO9000,focalloss,RefineDet,FCOS}. As shown in Table~\ref{tab:exp_1x}, when using ResNet-50 or ResNet-50-DC5 as backbones, Faster R-CNN~\cite{FasterRCNN} is able to achieve relatively satisfactory detection accuracy, while the original DETR~\cite{DETR} is still heavily under-trained. Other recently proposed DETR-based object detectors~\cite{DeformableDETR,ConditionalDETR,SMCA-DETR,AnchorDETR,DABDETR} obtain clearly better results over the original DETR~\cite{DETR}, but still have large gaps with Faster R-CNN~\cite{FasterRCNN}. With the proposed semantic-aligned matching mechanism incorporated into DETR~\cite{DETR}, the standalone \verb|SAM-DETR| without multi-scale feature fusion obtains significant performance gain over the original DETR baseline~\cite{DETR} (+11.9\%\,AP\;w/\,R50 and +13.2\%\,AP\;w/\,R50-DC5), achieves comparable detection accuracy as Faster R-CNN~\cite{FasterRCNN}, and also significantly outperforms other Transformer-based object detectors~\cite{DeformableDETR,ConditionalDETR,SMCA-DETR,AnchorDETR,DABDETR}.
Furthermore, since \verb|SAM-DETR| does not modify the attention mechanism nor does it modify the training strategy, one unique advantage of \verb|SAM-DETR| is that it can be easily integrated with other DETR's convergence solutions. As shown in Table\;\ref{tab:exp_1x}, our approach can be integrated with SMCA-DETR~\cite{SMCA-DETR}, DN-DETR~\cite{DN-DETR}, or both of them. The integrated methods achieve consistent and significant performance gain over the standalone \verb|SAM-DETR| as well as their respective baselines~\cite{SMCA-DETR,DN-DETR}. Our methods even significantly outperform the known-to-be-fast-converging Faster R-CNN~\cite{FasterRCNN} by very large margins. These results verify the effectiveness of our proposed semantic-aligned matching mechanism and its strong compatibility.

In Table\;\ref{tab:exp_1x}, we also present the object detection performance of our proposed \verb|SAM-DETR++| that is extended to fuse multi-scale features, as well as the performance of other detectors exploiting multi-scale features. As shown in Table\;\ref{tab:exp_1x}, fusing multi-scale features via the introduced semantic-aligned matching mechanism significantly improves the detection accuracy over \verb|SAM-DETR-R50-DC5| with even reduced computational cost. In addition, it is noteworthy that \verb|SAM-DETR++-R50 w/ SMCA+DN| achieves a state-of-the-art detection performance of 44.8\%\,AP with only 12 training epochs, which outperforms the original \verb|DETR-R50-DC5|~\cite{DETR} trained for 500 epochs (43.3\%\,AP), reducing the required number of training epochs by more than 97.6\%. These results show that fusing multi-scale features via our proposed semantic-aligned matching mechanism further improves the detection accuracy and accelerates the convergence to a greater extent.

\begin{figure*}[t!]%
    \centering
    \vspace{-2.5mm}
    \subfloat
    {{\includegraphics[width=.44\linewidth]{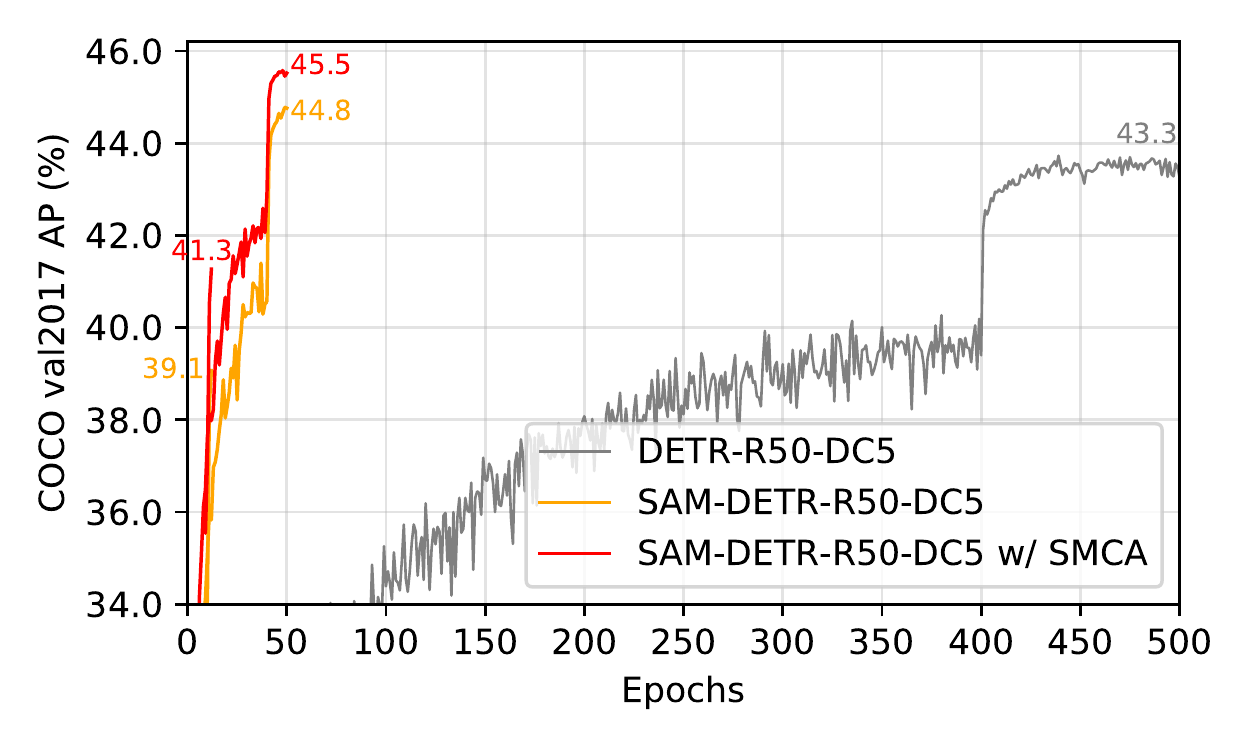} }}\;\;\;\;\;\;\;\;\;\;\;%
    \subfloat
    {{\includegraphics[width=.44\linewidth]{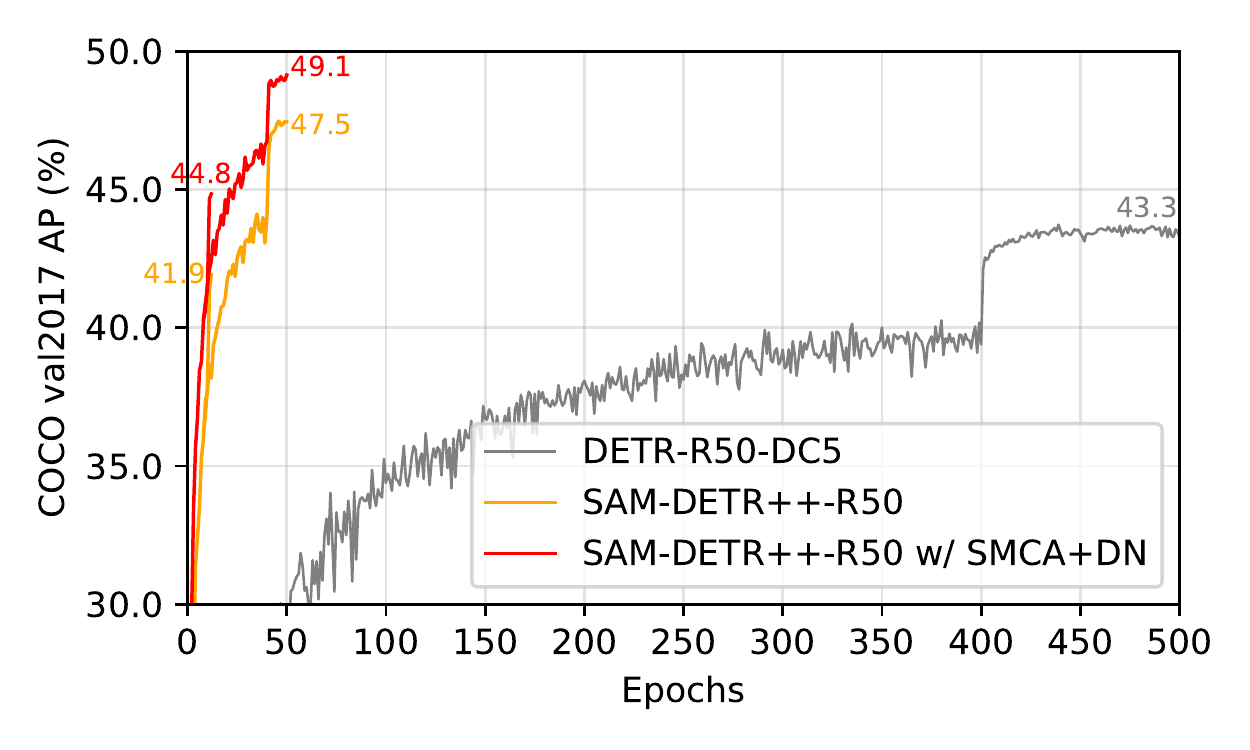} }}%
    \vspace{-2.5mm}
    \caption{The convergence curves of DETR, SAM-DETR (\textit{w/o} multi-scale feature fusion), and SAM-DETR++ (\textit{w/} multi-scale feature fusion). DETR is trained with 500 epochs, with the learning rate dropped at the 400$^{\text{th}}$ epoch. SAM-DETR and SAM-DETR++ are trained under the 12-epoch (1x) and 50-epoch learning schedules. Our methods converge much faster and achieve clearly better detection performance over the original DETR.}%
    \label{fig:convergence_curves}%
    \vspace{+1.0mm}
\end{figure*}

\vspace{+2mm}
\noindent
\textbf{Results under Longer Training Schedules.\;\;\;}
Table\;\ref{tab:exp_long} further compares \verb+SAM-DETR+++ with other state-of-the-art object detectors under the longer training schedules. When trained for 50 epochs, our proposed \verb+SAM-DETR+++ already outperforms the original DETR~\cite{DETR} trained for 500 epochs by large margins, and also achieves state-of-the-art performance among all Transformer-based object detectors. In addition, as \verb+SAM-DETR+++ works from a distinct perspective from existing solutions, combining our proposed \verb+SAM-DETR+++ with SMCA~\cite{SMCA-DETR} and DN~\cite{DN-DETR} (\verb|SAM-DETR++ w/ SMCA| \,and\, \verb|SAM-DETR++ w/ SMCA+DN|) brings further performance gains, achieving the state-of-the-art accuracy of 49.1\%\,AP on COCO val\,2017 with ResNet-50, without bells and whistles.

\vspace{+2.0mm}
\noindent
\textbf{Convergence Curves.\;\;\;}
We also present the convergence curves of the proposed \verb|SAM-DETR| and \verb+SAM-DETR+++ in Fig.~\ref{fig:convergence_curves}, which show significantly accelerated convergence speed of our methods over the baselines. These experiment results well validate our method's superior learning efficiency, good detection accuracy, and strong compatibility.

\newcolumntype{?}{!{\vrule width 1.25pt}}
\begin{table*}[t!]
\begin{center}
\caption{Ablation study on the design choices of SAM-DETR++. Results are obtained on COCO val\,2017 under the 12-epoch (1x) learning schedule.}
\label{tab:samdetr_ablation1}
\vspace{-8.0mm}
\centering
\setlength{\tabcolsep}{3.888pt}
\rowcolors{4}{gray!18}{white}
\resizebox{1.0\textwidth}{!}{
\begin{tabular}[t]{c|c|c|c|c|c|c|c?cccccc}
\toprule[1.35pt]

Semantics & \multicolumn{4}{c|}{\scriptsize Query Re-Sampling Strategy} & Feature & Remove & Multi-Scale & \multirow{2}{*}{AP} & \multirow{2}{*}{AP$_{\rm 50}$} & \multirow{2}{*}{AP$_{\rm 75}$} & \multirow{2}{*}{AP$_{\rm S}$} & \multirow{2}{*}{AP$_{\rm M}$} & \multirow{2}{*}{AP$_{\rm L}$} \\ 

\cline{2-5}

Aligner & \scriptsize AvgPool & \scriptsize MaxPool & \scriptsize Keypoint\,x1 & \scriptsize Keypoints\,x8 & Reweighting & Dropout & Feature Fusion & & \\

\midrule[1.1pt]

& & & & & & & & 22.3 & 39.5 & 22.2 & 6.6 & 22.8 & 36.6 \\

$\checkmark$ & $\checkmark$ & & & & & & & 25.2 & 48.9 & 23.3 & 8.9 & 26.4 & 41.3  \\

$\checkmark$ & & $\checkmark$ & & & & & & 27.0 & 50.2 & 25.8 & 10.3 & 28.0 & 43.9 \\

$\checkmark$ & & & $\checkmark$ & & & & & 28.6 & 50.3 & 28.1 & 12.4 & 31.2 & 44.4 \\

$\checkmark$ & & & $\checkmark$ & & $\checkmark$ & & & 30.3 & 52.0 & 29.8 & 12.4 & 32.8 & 47.3 \\

$\checkmark$ & & & & $\checkmark$ & & & & 32.0 & 53.4 & 32.8 & 13.5 & 35.3 & 49.2 \\

$\checkmark$ & & & & $\checkmark$ & $\checkmark$ & & & 33.1 & 54.2 & 33.7 & 13.9 & 36.5 & 51.7 \\

$\checkmark$ & & & & $\checkmark$ & $\checkmark$ & $\checkmark$ & & 34.2 & 55.8 & 35.3 & 15.0 & 37.7 & 52.5 \\

& & & & & & $\checkmark$ & $\checkmark$ & 29.1 & 53.2 & 28.6 & 11.5 & 31.8 & 46.0  \\

$\checkmark$ & & & & $\checkmark$ & $\checkmark$ & $\checkmark$ & $\checkmark$ & \textbf{41.9} & \textbf{60.5} & \textbf{45.3} & \textbf{24.6} & \textbf{45.5} & \textbf{57.4} \\

\bottomrule[1.35pt]

\end{tabular}
}
\end{center}
\end{table*}

\subsection{Ablation Study} \label{sec:ablation_study}

We conduct comprehensive ablation experiments to validate the effectiveness of our designs in SAM-DETR++. The ablation experiments are performed with ResNet-50~\cite{resnet} as the backbone network and under the short 12-epoch (1x) learning schedule.

\vspace{+2.0mm}
\noindent
\textbf{Effect of Semantic-Aligned Matching.\;\;\;}
The first row in Table\;\ref{tab:samdetr_ablation1} shows the detection result of the origianl DETR baseline~\cite{DETR}. As shown in Table\;\ref{tab:samdetr_ablation1}, the proposed Semantic Aligner, together with any query re-sampling strategy, consistently improves the performance over the baseline. We highlight that even with the naive max-pooling re-sampling, AP and AP$_{\rm 50}$ significantly improves by 4.7\% and 10.7\%, respectively. The results validate our claim that the proposed semantic-aligned matching mechanism effectively eases the matching difficulty between object queries and their corresponding target features, thus accelerating the training convergence of DETR.

\vspace{+2.0mm}
\noindent
\textbf{Effect of Semantic-Aligned Matching with Representative Keypoints.\;\;\;}
As shown in Table\;\ref{tab:samdetr_ablation1}, different object query re-sampling strategies lead to large variance in detection performance. Max-pooling performs clearly better than average-pooling, which suggests that object detection relies more on salient features rather than treating all features within reference boxes equally. This motivates us to explicitly search representative keypoints and employ their features for the introduced semantic-aligned matching mechanism. Results show that searching just one keypoint and re-sampling its features as new object queries outperforms the naive re-sampling strategies (AvgPool and MaxPool). Furthermore, searching multiple representative keypoints can naturally work with the multi-head attention mechanism~\cite{transformer} to further strengthen the representation capability of the re-sampled new object queries produced by Semantics Aligner and achieve superior performance.

\begin{figure}[t] 
\begin{center}
   \includegraphics[width=0.725\linewidth]{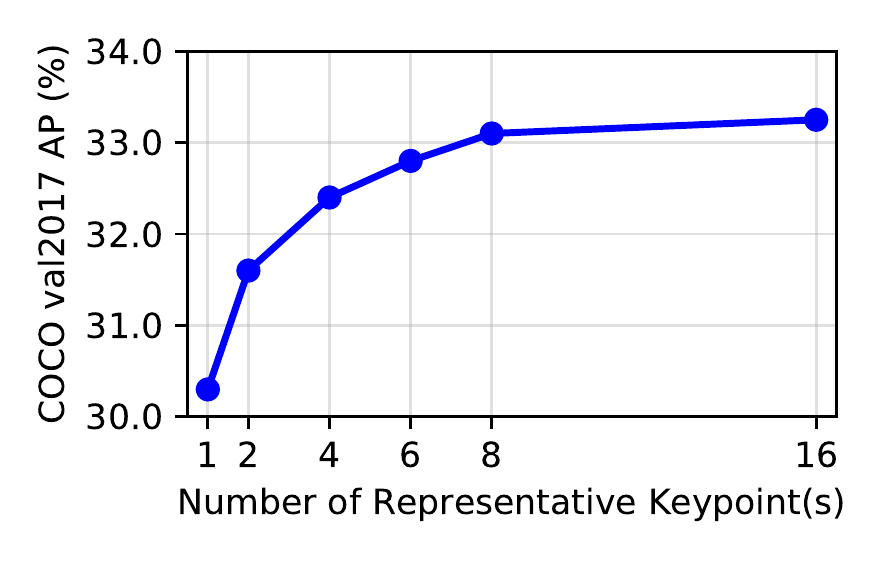}
\end{center}
\vspace{-8mm}
\caption{
Ablation study on the number of searched representative keypoint(s) for each object query. Results are obtained on COCO val\,2017 under the 12-epoch (1x) learning schedule without multi-scale feature fusion and removing dropout.
}
\label{fig:SAMDETR_ablation_num_sp}
\vspace{+1.25mm}
\end{figure}

Fig.\,\ref{fig:SAMDETR_ablation_num_sp} also studies the effect of the number of representative keypoints for each object query. As the figure shows, the performance increases as the number of keypoints increases and saturates at 8. Therefore, we set the number of representative keypoints at 8 by default in our approach.

\vspace{+2.5mm}
\noindent
\textbf{Searching within Reference Boxes \textit{vs.} Searching within Images.\;\;\;}
As introduced in Section\;\ref{sec:SAM_DETR_semantic_aligner}, the representative keypoints are searched within their corresponding reference boxes. We also evaluate the performance when allowing representative keypoints to be outside their corresponding reference boxes by relaxing the search range constraint. As shown in Table\;\ref{tab:samdetr_ablation2}, searching representative keypoints at the image scale impedes the performance. We suspect the performance drop is due to the increased difficulty of matching given a larger search range. Note that, in the original DETR~\cite{DETR}, object queries do not have explicit search ranges, which corresponds to the setup of image-scale searching. While our proposed SAM-DETR++ models learnable reference boxes with interpretable meanings, which can effectively narrow down the search range, leading to faster training convergence.

\begin{table}[t]
\begin{center}
\caption{Ablation study on the search range of representative keypoints. Results are obtained on COCO val\,2017 under the 12-epoch (1x) learning schedule without multi-scale feature fusion and removing dropout.}
\label{tab:samdetr_ablation2}
\vspace{-5.0mm}
\centering
\setlength{\tabcolsep}{6pt}
\resizebox{0.445\textwidth}{!}{
\begin{tabular}[t]{c|c?ccc}
\toprule[1.2333pt]

\multicolumn{2}{c?}{\scriptsize Keypoint Search Range} & \multirow{2}{*}{AP} & \multirow{2}{*}{AP$_{\rm 0.5}$} & \multirow{2}{*}{AP$_{\rm 0.75}$} \\ 

\cline{1-2}

\scriptsize within ref box & \scriptsize within image & & & \\

\midrule[0.8pt]

$\checkmark$ & & 33.1 & 54.2 & 33.7 \\

& $\checkmark$ & 30.0 & 52.3 & 29.2 \\

\bottomrule[1.2333pt]

\end{tabular}
}
\end{center}
\end{table}

\vspace{+2.5mm}
\noindent
\textbf{Effect of Feature Reweighting with Previous Query Embeddings.\;\;\;}
As discussed in Section\;\ref{sec:SAM_DETR_semantic_aligner}, previous object queries' embeddings contain helpful information for object detection, which cannot be directly leveraged due to the introduced re-sampling process. As a workaround, we propose to perform feature reweighting on re-sampled query embeddings based on previous query embeddings. This incorporates information from previous query embeddings while does not impede the aligned semantics. As shown in Table\;\ref{tab:samdetr_ablation1}, the proposed feature reweighting mechanism consistently boosts performance, indicating its effectiveness.

\vspace{+2.5mm}
\noindent
\textbf{Effect of Removing Dropout in Transformer.\;\;\;}
As shown in Table\;\ref{tab:samdetr_ablation1}, the simple tweak of removing dropout~\cite{srivastava2014dropout} in Transformer~\cite{transformer} for object detection increases the performance of \verb|SAM-DETR| without involving multi-scale feature fusion by 1.1\%\,AP, at no extra computational cost. 


\vspace{+2.5mm}
\noindent
\textbf{Effect of Multi-Scale Feature Fusion with Aligned Semantics.\;\;\;}
As shown in Table\;\ref{tab:samdetr_ablation1}, on top of \verb|SAM-DETR|, incorporating multi-scale feature fusion improves the detection performance by a considerable margin of 7.7\%\,AP. This verifies that multi-scale feature fusion effectively reduces the complexity of representing objects of different sizes and can adaptively choose appropriate feature scales for object representation, leading to further performance gain.

It is noteworthy that multi-scale feature fusion highly depends on our proposed semantic-aligned matching mechanism. As shown in Table\;\ref{tab:samdetr_ablation1}, performing multi-scale feature fusion without our proposed semantic-aligned matching only yields a poor performance of 29.1\%\,AP (-12.8\%\,AP compared with \verb|SAM-DETR++|). This is because there exists inevitable unaligned semantics across different feature scales. Without imposing aligned semantics, directly fusing multi-scale features that are projected into different feature embedding spaces (\textit{i.e.}, with unaligned semantics) causes extra matching difficulty in cross-attention, as explained in Section\;\ref{sec:introduction} and Fig.\,\ref{fig:fig1}.

\subsection{Further Discussions}

\vspace{+0.6mm}
\noindent
\textbf{On the Compatibility among SAM-DETR++, SMCA-DETR~\cite{SMCA-DETR}, and DN-DETR~\cite{DN-DETR}.\;\;\;}
One of the key advantages of the proposed SAM-DETR++ is its excellent compatibility, which we demonstrate by integrating it with SMCA-DETR~\cite{SMCA-DETR} and DN-DETR~\cite{DN-DETR} and achieving superior detection performance. The reason behind their excellent compatibility is that each of them effectively accelerates the convergence of DETR from distinct perspectives, thus complementing each other. Concretely, SMCA-DETR~\cite{SMCA-DETR} accelerates DETR's convergence by imposing strong spatial constraints for the cross-attention module, in which each object query is limited to attend to a specific region adaptively. SMCA-DETR~\cite{SMCA-DETR} effectively reduces the search space for each object query in cross-attention, thus improving the training convergence. DN-DETR~\cite{DN-DETR} proposes a de-noising training strategy to mitigate the instability of bipartite graph matching that causes inconsistent optimization goals in DETR's early training stages. With its proposed de-noising training strategy, the optimization objectives of DETR become consistent even in early training stages, which accelerates DETR's training convergence. Unlike the above two methods, our proposed SAM-DETR++ aims to reduce the matching difficulty between object query and encoded image features by enforcing aligned semantics, which encourages each object query to attend to those features with similar semantics. We demonstrate that with adequately addressed factors that obstruct convergence, Transformer-based detectors do not fall behind conventional ConvNet-based detectors~\cite{FasterRCNN,FCOS,efficientdet} in terms of convergence speed, with even superior performance and simpler pipelines.

\vspace{+2.0mm}
\noindent
\textbf{Relevance and Difference with Sparse R-CNN~\cite{SparseRCNN}.\;\;\;}
We encode instance-level information with reference boxes and object queries, which have certain similarities to the proposal boxes and proposal features in Sparse R-CNN~\cite{SparseRCNN}. Besides, both methods leverage RoIAlign~\cite{MaskRCNN} to pool region features. However, the two methods are fundamentally distinct. As a member of the R-CNN family, Sparse R-CNN directly feeds the pooled features to a heavy R-CNN head to produce region-wise detection results. In contrast, our proposed SAM-DETR++ searches and extracts objects' salient features from the pooled region features using a lightweight network. The extracted features are fed to the Transformer modules for global predictions with accelerated convergence.

\vspace{+2.0mm}
\noindent
\textbf{Are Learned Reference Boxes Sensitive to Gaps across Datasets?\;\;\;}
The learned reference boxes encode statistical information for object distribution of specific datasets. To study whether these reference boxes affect generalization across datasets, we train and evaluate \verb|SAM-DETR++ w/ SMCA| on Pascal VOC~\cite{PascalVOC} (with notably different statistics from COCO~\cite{MSCOCO}) over three setups for reference boxes: \textit{(i)} learning from scratch on Pascal VOC, \textit{(ii)} inheriting from COCO and remaining fixed, and \textit{(iii)} remaining fixed from random initialization. Except for the reference boxes, all other parameters are trained on Pascal VOC. We use Pascal VOC trainval\,07+12 for training and use test\,07 for evaluation. Results in Table\;\ref{tab:SAMDETR_gap_across_datasets} show that the learned reference boxes generalize well across datasets. Even with totally random reference boxes, SAM-DETR++ can still deliver satisfactory detection accuracy. This is because \textit{(i)} the reference boxes are dense enough to cover most image regions, and \textit{(ii)} SAM-DETR++ involves multiple stages of box adjustment, thus initial reference boxes do not have a clear impact on final predictions.

\begin{table}[t]
    \centering
    \caption{Learnable reference boxes in SAM-DETR++ are not sentitive to gaps across datasets.}
    \label{tab:SAMDETR_gap_across_datasets}
    \vspace{-2.0mm}
    \setlength{\tabcolsep}{5pt}
    \resizebox{0.485\textwidth}{!}{
    \begin{tabular}{c|c|ccc}
        \toprule[1.2pt]
        \multirow{2}{*}{Ref.\,Box} & initialization & scratch & pretrained on COCO & random \\ \cline{2-5}
         & trainable\,? & $\checkmark$ & $\times$ & $\times$ \\
        \midrule[1.0pt]
        \multicolumn{2}{c|}{mAP\,@\,0.5 (\%)} & 79.6 & 79.5 & 79.3 \\
        \bottomrule[1.2pt]
    \end{tabular}
    }
    \vspace{+3.5mm}
\end{table}

\section{Conclusion}   \label{sec:SAMDETR_conclusion}

This paper presents SAM-DETR++ to accelerate the convergence of DETR. The core of SAM-DETR++ is a plug-and-play module that semantically aligns object queries and encoded image features to facilitate the matching procedure between them. It also explicitly searches multiple representative keypoints with the most discriminative features for semantic-aligned matching. Besides, on the basis of semantic-aligned matching, SAM-DETR++ can further benefit from multi-scale feature fusion in a coarse-to-fine manner. By simply introducing a plug-and-play module, our proposed SAM-DETR++ accelerates DETR's convergence from a unique perspective, and thus can be easily integrated with existing convergence solutions to boost performance to a greater extent. On the COCO benchmark, the fully-fledged SAM-DETR++ achieves 44.8\%\,AP with only 12 training epochs, outperforming Faster R-CNN by a large margin. It also achieves state-of-the-art detection accuracy among Transformer-based detectors. We hope our work paves the way for more comprehensive research and applications of Transformer-based object detectors.

\ifCLASSOPTIONcompsoc
  \section*{Acknowledgments}
\else
  \section*{Acknowledgment}
\fi

This study is supported under the RIE\,2020 Industry Alignment Fund – Industry Collaboration Projects (IAF-ICP) Funding Initiative, as well as cash and in-kind contribution from the industry partner(s).

\ifCLASSOPTIONcaptionsoff
  \newpage
\fi



\bibliographystyle{IEEEtran}
\bibliography{bib.bib}

\begin{thebibliography}{10}
\providecommand{\url}[1]{#1}
\csname url@samestyle\endcsname
\providecommand{\newblock}{\relax}
\providecommand{\bibinfo}[2]{#2}
\providecommand{\BIBentrySTDinterwordspacing}{\spaceskip=0pt\relax}
\providecommand{\BIBentryALTinterwordstretchfactor}{4}
\providecommand{\BIBentryALTinterwordspacing}{\spaceskip=\fontdimen2\font plus
\BIBentryALTinterwordstretchfactor\fontdimen3\font minus
  \fontdimen4\font\relax}
\providecommand{\BIBforeignlanguage}[2]{{%
\expandafter\ifx\csname l@#1\endcsname\relax
\typeout{** WARNING: IEEEtran.bst: No hyphenation pattern has been}%
\typeout{** loaded for the language `#1'. Using the pattern for}%
\typeout{** the default language instead.}%
\else
\language=\csname l@#1\endcsname
\fi
#2}}
\providecommand{\BIBdecl}{\relax}
\BIBdecl

\bibitem{Liu2019DeepLF}
L.~Liu, W.~Ouyang, X.~Wang, P.~Fieguth, J.~Chen, X.~Liu, and
  M.~Pietik{\"a}inen, ``Deep learning for generic object detection: A survey,''
  \emph{International Journal of Computer Vision}, vol. 128, pp. 261--318,
  2020.

\bibitem{FasterRCNN}
S.~Ren, K.~He, R.~B. Girshick, and J.~Sun, ``Faster {R-CNN}: Towards real-time
  object detection with region proposal networks,'' \emph{IEEE Transactions on
  Pattern Analysis and Machine Intelligence}, vol.~39, pp. 1137--1149, 2015.

\bibitem{YOLO9000}
J.~Redmon and A.~Farhadi, ``{YOLO 9000}: Better, faster, stronger,'' in
  \emph{CVPR}, 2017.

\bibitem{FCOS}
Z.~Tian, C.~Shen, H.~Chen, and T.~He, ``{FCOS}: Fully convolutional one-stage
  object detection,'' in \emph{ICCV}, 2019.

\bibitem{DETR}
N.~Carion, F.~Massa, G.~Synnaeve, N.~Usunier, A.~Kirillov, and S.~Zagoruyko,
  ``End-to-end object detection with transformers,'' in \emph{ECCV}, 2020.

\bibitem{transformer}
A.~Vaswani, N.~Shazeer, N.~Parmar, J.~Uszkoreit, L.~Jones, A.~N. Gomez,
  L.~Kaiser, and I.~Polosukhin, ``Attention is all you need,'' in
  \emph{NeurIPS}, 2017.

\bibitem{MSCOCO}
T.-Y. Lin, M.~Maire, S.~J. Belongie, L.~D. Bourdev, R.~B. Girshick, J.~Hays,
  P.~Perona, D.~Ramanan, P.~Doll{\'a}r, and C.~L. Zitnick, ``Microsoft {COCO}:
  Common objects in context,'' in \emph{ECCV}, 2014.

\bibitem{focalloss}
T.-Y. Lin, P.~Goyal, R.~Girshick, K.~He, and P.~Doll{\'a}r, ``Focal loss for
  dense object detection,'' in \emph{ICCV}, 2017.

\bibitem{DeformableDETR}
X.~Zhu, W.~Su, L.~Lu, B.~Li, X.~Wang, and J.~Dai, ``{Deformable DETR}:
  Deformable transformers for end-to-end object detection,'' in \emph{ICLR},
  2021.

\bibitem{ConditionalDETR}
D.~Meng, X.~Chen, Z.~Fan, G.~Zeng, H.~Li, Y.~Yuan, L.~Sun, and J.~Wang,
  ``Conditional {DETR} for fast training convergence,'' in \emph{ICCV}, 2021.

\bibitem{SMCA-DETR}
P.~Gao, M.~Zheng, X.~Wang, J.~Dai, and H.~Li, ``Fast convergence of {DETR} with
  spatially modulated co-attention,'' in \emph{ICCV}, 2021.

\bibitem{AnchorDETR}
Y.~Wang, X.~Zhang, T.~Yang, and J.~Sun, ``{Anchor DETR}: Query design for
  {Transformer}-based detector,'' in \emph{AAAI}, 2022.

\bibitem{SAM-DETR}
G.~Zhang, Z.~Luo, Y.~Yu, K.~Cui, and S.~Lu, ``Accelerating {DETR} convergence
  via semantic-aligned matching,'' in \emph{CVPR}, 2022.

\bibitem{Siam-FC}
L.~Bertinetto, J.~Valmadre, J.~F. Henriques, A.~Vedaldi, and P.~H. Torr,
  ``Fully-convolutional siamese networks for object tracking,'' in \emph{ECCV},
  2016.

\bibitem{SiamRPN}
B.~Li, J.~Yan, W.~Wu, Z.~Zhu, and X.~Hu, ``High performance visual tracking
  with siamese region proposal network,'' in \emph{CVPR}, 2018.

\bibitem{SiamRPN++}
B.~Li, W.~Wu, Q.~Wang, F.~Zhang, J.~Xing, and J.~Yan, ``{SiamRPN++}: Evolution
  of siamese visual tracking with very deep networks,'' in \emph{CVPR}, 2019.

\bibitem{SiamRCNN}
P.~Voigtlaender, J.~Luiten, P.~H. Torr, and B.~Leibe, ``Siam {R-CNN}: Visual
  tracking by re-detection,'' in \emph{CVPR}, 2020.

\bibitem{TransformerTrack}
N.~Wang, W.~Zhou, J.~Wang, and H.~Li, ``Transformer meets tracker: Exploiting
  temporal context for robust visual tracking,'' in \emph{CVPR}, 2021.

\bibitem{TransT}
X.~Chen, B.~Yan, J.~Zhu, D.~Wang, X.~Yang, and H.~Lu, ``Transformer tracking,''
  in \emph{CVPR}, 2021.

\bibitem{chung2017two}
D.~Chung, K.~Tahboub, and E.~J. Delp, ``A two stream siamese convolutional
  neural network for person re-identification,'' in \emph{ICCV}, 2017.

\bibitem{zheng2019re}
M.~Zheng, S.~Karanam, Z.~Wu, and R.~J. Radke, ``Re-identification with
  consistent attentive siamese networks,'' in \emph{CVPR}, 2019.

\bibitem{wu2018and}
L.~Wu, Y.~Wang, J.~Gao, and X.~Li, ``Where-and-when to look: Deep siamese
  attention networks for video-based person re-identification,'' \emph{IEEE
  Transactions on Multimedia}, vol.~21, no.~6, pp. 1412--1424, 2018.

\bibitem{shen2017deep}
C.~Shen, Z.~Jin, Y.~Zhao, Z.~Fu, R.~Jiang, Y.~Chen, and X.-S. Hua, ``Deep
  siamese network with multi-level similarity perception for person
  re-identification,'' in \emph{ACM MM}, 2017.

\bibitem{Shen_2017_ICCV}
Y.~Shen, T.~Xiao, H.~Li, S.~Yi, and X.~Wang, ``Learning deep neural networks
  for vehicle {Re-ID} with visual-spatio-temporal path proposals,'' in
  \emph{ICCV}, 2017.

\bibitem{SiameseOneshotImageRecognition}
G.~Koch, R.~Zemel, and R.~Salakhutdinov, ``Siamese neural networks for one-shot
  image recognition,'' in \emph{ICML {D}eep {L}earning {W}orkshop}, 2015.

\bibitem{ProtoNet}
J.~Snell, K.~Swersky, and R.~Zemel, ``Prototypical networks for few-shot
  learning,'' in \emph{NeurIPS}, 2017.

\bibitem{RelationNetwork}
F.~Sung, Y.~Yang, L.~Zhang, T.~Xiang, P.~H. Torr, and T.~M. Hospedales,
  ``Learning to compare: Relation network for few-shot learning,'' in
  \emph{CVPR}, 2018.

\bibitem{NEURIPS2019_92af93f7}
T.-I. Hsieh, Y.-C. Lo, H.-T. Chen, and T.-L. Liu, ``One-shot object detection
  with co-attention and co-excitation,'' in \emph{NeurIPS}, 2019.

\bibitem{MetaDETR}
G.~Zhang, Z.~Luo, K.~Cui, and S.~Lu, ``{Meta-DETR}: Image-level few-shot object
  detection with inter-class correlation exploitation,'' \emph{arXiv preprint
  arXiv:2103.11731}, 2021.

\bibitem{DN-DETR}
F.~Li, H.~Zhang, S.~Liu, J.~Guo, L.~M. Ni, and L.~Zhang, ``{DN-DETR}:
  Accelerate {DETR} training by introducing query denoising,'' in \emph{CVPR},
  2022.

\bibitem{CascadeRCNN}
Z.~Cai and N.~Vasconcelos, ``{Cascade R-CNN}: High quality object detection and
  instance segmentation,'' \emph{IEEE Transactions on Pattern Analysis and
  Machine Intelligence}, vol.~43, pp. 1483--1498, 2021.

\bibitem{LibraRCNN}
J.~Pang, K.~Chen, J.~Shi, H.~Feng, W.~Ouyang, and D.~Lin, ``Libra {R-CNN}:
  Towards balanced learning for object detection,'' in \emph{CVPR}, 2019.

\bibitem{tychsen2018improving}
L.~Tychsen-Smith and L.~Petersson, ``Improving object localization with fitness
  {NMS} and bounded {IoU} loss,'' in \emph{CVPR}, 2018.

\bibitem{RelationNetworkObjectDetection}
H.~Hu, J.~Gu, Z.~Zhang, J.~Dai, and Y.~Wei, ``Relation networks for object
  detection,'' in \emph{CVPR}, 2018.

\bibitem{CADNet}
G.~Zhang, S.~Lu, and W.~Zhang, ``{CAD-Net}: A context-aware detection network
  for objects in remote sensing imagery,'' \emph{IEEE Transactions on
  Geoscience and Remote Sensing}, vol.~57, no.~12, pp. 10\,015--10\,024, 2019.

\bibitem{metarcnn}
X.~Yan, Z.~Chen, A.~Xu, X.~Wang, X.~Liang, and L.~Lin, ``{Meta R-CNN}: Towards
  general solver for instance-level low-shot learning,'' in \emph{ICCV}, 2019.

\bibitem{masktextspotter}
M.~Liao, P.~Lyu, M.~He, C.~Yao, W.~Wu, and X.~Bai, ``{Mask TextSpotter}: An
  end-to-end trainable neural network for spotting text with arbitrary
  shapes,'' \emph{IEEE Transactions on Pattern Analysis and Machine
  Intelligence}, vol.~43, no.~2, pp. 532--548, 2021.

\bibitem{FSDetView}
Y.~Xiao and R.~Marlet, ``Few-shot object detection and viewpoint estimation for
  objects in the wild,'' in \emph{ECCV}, 2020.

\bibitem{fsod}
Q.~Fan, W.~Zhuo, C.-K. Tang, and Y.-W. Tai, ``Few-shot object detection with
  attention-{RPN} and multi-relation detector,'' in \emph{CVPR}, 2020.

\bibitem{fsdet}
X.~Wang, T.~E. Huang, T.~Darrell, J.~E. Gonzalez, and F.~Yu, ``Frustratingly
  simple few-shot object detection,'' in \emph{ICML}, 2020.

\bibitem{SSD}
W.~Liu, D.~Anguelov, D.~Erhan, C.~Szegedy, S.~Reed, C.-Y. Fu, and A.~C. Berg,
  ``{SSD}: Single shot multibox detector,'' in \emph{ECCV}, 2016.

\bibitem{FewshotReweighting}
B.~Kang, Z.~Liu, X.~Wang, F.~Yu, J.~Feng, and T.~Darrell, ``Few-shot object
  detection via feature reweighting,'' in \emph{ICCV}, 2019.

\bibitem{RefineDet}
S.~Zhang, L.~Wen, X.~Bian, Z.~Lei, and S.~Z. Li, ``Single-shot refinement
  neural network for object detection,'' in \emph{CVPR}, 2018.

\bibitem{RFBNet}
S.~Liu, D.~Huang, and Y.~Wang, ``Receptive field block net for accurate and
  fast object detection,'' in \emph{ECCV}, 2018.

\bibitem{m2det}
Q.~Zhao, T.~Sheng, Y.~Wang, Z.~Tang, Y.~Chen, L.~Cai, and H.~Ling, ``{M2Det}: A
  single-shot object detector based on multi-level feature pyramid network,''
  in \emph{AAAI}, 2019.

\bibitem{efficientdet}
M.~Tan, R.~Pang, and Q.~V. Le, ``{EfficientDet}: Scalable and efficient object
  detection,'' in \emph{CVPR}, 2020.

\bibitem{zhou2019objects}
X.~Zhou, D.~Wang, and P.~Kr{\"a}henb{\"u}hl, ``Objects as points,'' in
  \emph{arXiv preprint arXiv:1904.07850}, 2019.

\bibitem{incrementalfsdet}
J.-M. Perez-Rua, X.~Zhu, T.~M. Hospedales, and T.~Xiang, ``Incremental few-shot
  object detection,'' in \emph{CVPR}, 2020.

\bibitem{PNPDet}
G.~Zhang, K.~Cui, R.~Wu, S.~Lu, and Y.~Tian, ``{PNPDet}: Efficient few-shot
  detection without forgetting via plug-and-play sub-networks,'' in
  \emph{WACV}, 2021.

\bibitem{ExtremeNet}
X.~Zhou, J.~Zhuo, and P.~Krahenbuhl, ``Bottom-up object detection by grouping
  extreme and center points,'' in \emph{CVPR}, 2019.

\bibitem{EfficientDETR}
Z.~Yao, J.~Ai, B.~Li, and C.~Zhang, ``Efficient {DETR}: improving end-to-end
  object detector with dense prior,'' \emph{arXiv preprint arXiv:2104.01318},
  2021.

\bibitem{SparseDETR}
B.~Roh, J.~Shin, W.~Shin, and S.~Kim, ``Sparse {DETR}: Efficient end-to-end
  object detection with learnable sparsity,'' in \emph{ICLR}, 2022.

\bibitem{ViDT}
H.~Song, D.~Sun, S.~Chun, V.~Jampani, D.~Han, B.~Heo, W.~Kim, and M.-H. Yang,
  ``{ViDT}: An efficient and effective fully transformer-based object
  detector,'' in \emph{ICLR}, 2022.

\bibitem{PnPDETR}
T.~Wang, L.~Yuan, Y.~Chen, J.~Feng, and S.~Yan, ``{PnP-DETR}: Towards efficient
  visual analysis with {Transformers},'' in \emph{ICCV}, 2021.

\bibitem{DABDETR}
S.~Liu, F.~Li, H.~Zhang, X.~Yang, X.~Qi, H.~Su, J.~Zhu, and L.~Zhang,
  ``{DAB}-{DETR}: Dynamic anchor boxes are better queries for {DETR},'' in
  \emph{ICLR}, 2022.

\bibitem{FaceNet}
F.~Schroff, D.~Kalenichenko, and J.~Philbin, ``{FaceNet}: A unified embedding
  for face recognition and clustering,'' in \emph{CVPR}, 2015.

\bibitem{song2019occlusion}
L.~Song, D.~Gong, Z.~Li, C.~Liu, and W.~Liu, ``Occlusion robust face
  recognition based on mask learning with pairwise differential siamese
  network,'' in \emph{ICCV}, 2019.

\bibitem{tao2016siamese}
R.~Tao, E.~Gavves, and A.~W. Smeulders, ``Siamese instance search for
  tracking,'' in \emph{CVPR}, 2016.

\bibitem{dong2018triplet}
X.~Dong and J.~Shen, ``Triplet loss in siamese network for object tracking,''
  in \emph{ECCV}, 2018.

\bibitem{he2018twofold}
A.~He, C.~Luo, X.~Tian, and W.~Zeng, ``A twofold siamese network for real-time
  object tracking,'' in \emph{CVPR}, 2018.

\bibitem{zhu2018distractor}
Z.~Zhu, Q.~Wang, B.~Li, W.~Wu, J.~Yan, and W.~Hu, ``Distractor-aware siamese
  networks for visual object tracking,'' in \emph{ECCV}, 2018.

\bibitem{zhang2019deeper}
Z.~Zhang and H.~Peng, ``Deeper and wider siamese networks for real-time visual
  tracking,'' in \emph{CVPR}, 2019.

\bibitem{TransReID}
S.~He, H.~Luo, P.~Wang, F.~Wang, H.~Li, and W.~Jiang, ``{TransReID}:
  Transformer-based object re-identification,'' in \emph{ICCV}, 2021.

\bibitem{Meta-DETR_firstversion}
G.~Zhang, Z.~Luo, K.~Cui, and S.~Lu, ``{Meta-DETR}: Few-shot object detection
  via unified image-level meta-learning,'' \emph{arXiv preprint
  arXiv:2103.11731v2}, 2021.

\bibitem{DA-DETR}
J.~Zhang, J.~Huang, Z.~Luo, G.~Zhang, and S.~Lu, ``{DA-DETR}: Domain adaptive
  detection transformer by hybrid attention,'' \emph{arXiv preprint
  arXiv:2103.17084}, 2021.

\bibitem{CF-DETR}
X.~Cao, P.~Yuan, B.~Feng, and K.~Niu, ``{CF-DETR}: Coarse-to-fine transformers
  for end-to-end object detection,'' in \emph{AAAI}, 2022.

\bibitem{DynamicDETR}
X.~Dai, Y.~Chen, J.~Yang, P.~Zhang, L.~Yuan, and L.~Zhang, ``{Dynamic DETR}:
  End-to-end object detection with dynamic attention,'' in \emph{ICCV}, 2021.

\bibitem{DINO}
H.~Zhang, F.~Li, S.~Liu, L.~Zhang, H.~Su, J.~Zhu, L.~M. Ni, and H.-Y. Shum,
  ``{DINO}: {DETR} with improved denoising anchor boxes for end-to-end object
  detection,'' \emph{arXiv preprint arXiv:2203.03605}, 2022.

\bibitem{MaskRCNN}
K.~He, G.~Gkioxari, P.~Doll{\'a}r, and R.~Girshick, ``{Mask R-CNN},'' in
  \emph{ICCV}, 2017.

\bibitem{reppoints}
Z.~Yang, S.~Liu, H.~Hu, L.~Wang, and S.~Lin, ``{RepPoints}: Point set
  representation for object detection,'' in \emph{ICCV}, 2019.

\bibitem{wu2020cascade}
R.~Wu, G.~Zhang, S.~Lu, and T.~Chen, ``{Cascade EF-GAN}: Progressive facial
  expression editing with local focuses,'' in \emph{CVPR}, 2020.

\bibitem{reppointsv2}
Y.~Chen, Z.~Zhang, Y.~Cao, L.~Wang, S.~Lin, and H.~Hu, ``{RepPoints v2}:
  Verification meets regression for object detection,'' in \emph{NeurIPS},
  2020.

\bibitem{DefectGAN}
G.~Zhang, K.~Cui, T.-Y. Hung, and S.~Lu, ``{Defect-GAN}: High-fidelity defect
  synthesis for automated defect inspection,'' in \emph{WACV}, 2021.

\bibitem{Meta-DETR}
G.~Zhang, Z.~Luo, K.~Cui, and S.~Lu, ``{Meta-DETR}: Image-level few-shot object
  detection with inter-class correlation exploitation,'' \emph{arXiv preprint
  arXiv:2103.11731v3}, 2021.

\bibitem{FPN}
T.-Y. Lin, P.~Doll{\'a}r, R.~Girshick, K.~He, B.~Hariharan, and S.~Belongie,
  ``Feature pyramid networks for object detection,'' in \emph{CVPR}, 2017.

\bibitem{srivastava2014dropout}
N.~Srivastava, G.~Hinton, A.~Krizhevsky, I.~Sutskever, and R.~Salakhutdinov,
  ``Dropout: a simple way to prevent neural networks from overfitting,''
  \emph{The Journal of Machine Learning Research (JMLR)}, vol.~15, no.~1, pp.
  1929--1958, 2014.

\bibitem{SparseRCNN}
P.~Sun, R.~Zhang, Y.~Jiang, T.~Kong, C.~Xu, W.~Zhan, M.~Tomizuka, L.~Li,
  Z.~Yuan, C.~Wang, and P.~Luo, ``{Sparse R-CNN}: End-to-end object detection
  with learnable proposals,'' in \emph{CVPR}, 2021.

\bibitem{imagenet}
J.~Deng, W.~Dong, R.~Socher, L.-J. Li, K.~Li, and L.~Fei-Fei, ``{ImageNet}: A
  large-scale hierarchical image database,'' in \emph{CVPR}, 2009.

\bibitem{resnet}
K.~He, X.~Zhang, S.~Ren, and J.~Sun, ``Deep residual learning for image
  recognition,'' in \emph{CVPR}, 2016.

\bibitem{Adam}
D.~P. Kingma and J.~Ba, ``Adam: A method for stochastic optimization,'' in
  \emph{ICLR}, 2015.

\bibitem{AdamW}
I.~Loshchilov and F.~Hutter, ``Decoupled weight decay regularization,'' in
  \emph{ICLR}, 2019.

\bibitem{TSPRCNN}
Z.~Sun, S.~Cao, Y.~Yang, and K.~M. Kitani, ``Rethinking {Transformer}-based set
  prediction for object detection,'' in \emph{ICCV}, 2021.

\bibitem{up-detr}
Z.~Dai, B.~Cai, Y.~Lin, and J.~Chen, ``{UP-DETR}: Unsupervised pre-training for
  object detection with transformers,'' in \emph{CVPR}, 2021.

\bibitem{PascalVOC}
M.~Everingham, L.~Van~Gool, C.~K.~I. Williams, J.~Winn, and A.~Zisserman,
  ``{The Pascal Visual Object Classes {(VOC)} Challenge},'' \emph{International
  Journal of Computer Vision}, vol.~88, no.~2, pp. 303--338, 2010.

\end{thebibliography}
\end{document}